\documentclass{template/egpubl}
\JournalSubmission    
\CGFccby

\usepackage[T1]{fontenc}
\usepackage{template/dfadobe}  
\usepackage{adjustbox}

\usepackage{cite}  
\BibtexOrBiblatex

\electronicVersion
\PrintedOrElectronic
\ifpdf \pdfcompresslevel=9
\fi

\usepackage{template/egweblnk} 

\usepackage{pifont}
\usepackage{xcolor}
\usepackage{colortbl}
\usepackage{color}
\usepackage{tikz}
\usepackage{tabularx}
\usetikzlibrary{positioning, arrows.meta, shapes}
\usepackage{subcaption}
\usepackage{forest}
\usepackage{microtype}

\usepackage[cmex10]{amsmath}

\usepackage{cleveref}

\usepackage{array}

\usepackage{multirow}
\usepackage{booktabs}
\usepackage{enumitem}
\usepackage{amssymb}
\usepackage{lscape}
\usepackage{rotating}
\usepackage{pifont}
\usepackage{nicefrac}

\newcommand{\point}{p}				

\newcommand{\vertex}{\textbf{v}}		

\newcommand{\gaussians}{\mathcal{G}}

\newcommand{\neuralsurfacefunc}{f}  
\newcommand{\basicfunction}{\textbf{B}}
  
\newcommand{\params}{\theta}			

\newcommand{\featurevector}{\textbf{x}}   
\newcommand{\latentcode}{\featurevector}

\newcommand{\latentspace}{\mathcal{X}}	

\newcommand{\ie}{\emph{i.e., }}
\newcommand{\eg}{\emph{e.g., }}
\newcommand{\etal}{\emph{et al.}}

\newcommand{\noi}{\noindent}

\newcommand{\real}{\mathbb{R}}      	

\newcommand{\rthree}{\real^3}

\newcommand{\camera}{C}			

\newcommand{\polynomial}{\text{P}}

\newcommand{\tensor}{\mathcal{T}}

\newcommand{\outerproduct}{\circ}  

\newcommand{\domain}{\mathcal{D}}		
\newcommand{\geometryspace}{\mathcal{G}} %
\newcommand{\appearancespace}{\mathcal{A}} %

\newcommand{\stwo}{S^2}				
\newcommand{\domainpoint}{s}             
\newcommand{\inputparams}{\textbf{s}}

\newcommand{\First}{\textbf{First}}
\newcommand{\Second}{\textbf{Second}}

\def\cov{\Sigma}                     

\newcommand{\pose}{\theta}

\newcommand{\thecolor}{c}
\newcommand{\volumedensity}{\sigma}
\newcommand{\transmittance}{T}
\newcommand{\probabilitydensity}{\varphi}

\newcommand{\implicitfunc}{F}

\newcommand{\npoints}{N}

\newcommand{\viewdir}{\textbf{d}}
\newcommand{\ray}{r}

\newcommand{\transformation}{T}

\newcommand{\thetime}{t}

\newcommand{\shapeproperties}{\xi}

\newcommand{\neuralnetwork}{\text{NN}}

\newcommand{\MLP}{\text{MLP}}

\newcommand{\conditions}{\mathcal{C}}

\newcommand{\motionconditions}{\mathcal{C}_\pose}

\newcommand{\enc}{\text{enc}}

\usepackage[table]{xcolor}
\usepackage{tikz}

\definecolor{cbTeal}{RGB}{0, 114, 178}
\definecolor{cbOrange}{RGB}{230, 159, 0}
\definecolor{cbGray}{RGB}{153, 153, 153}

\definecolor{customOrange}{HTML}{FFD5AF}
\definecolor{customYellow}{HTML}{FFFDAB}
\definecolor{customRose}{HTML}{CC79A7}
\definecolor{customGreen}{HTML}{A9FFCD}
\definecolor{customBlue}{HTML}{D5B1FF}

\DeclareRobustCommand{\high}{\tikz[baseline=-0.8ex]\draw[cbTeal,fill=cbTeal,thick] (0,0) circle (1ex);}
\DeclareRobustCommand{\midlow}{\tikz[baseline=-0.8ex]{\draw[cbOrange,thick] (0,0) circle (1ex); \fill[cbOrange] (0,0) -- (90:1ex) arc (90:-90:1ex) -- cycle;}}
\DeclareRobustCommand{\low}{\tikz[baseline=-0.8ex]\draw[cbGray,thick] (0,0) circle (1ex);}

\title[Recent Advances and Trends in Learning-based 3D Representations]%
      {Recent Advances and Trends in Learning-based 3D Representations}

\author[Adrien Schockaert\& Hamid Laga \& Hazem Wannous \& Vincent Magnier \& Guillaume Dufaye \& Jean-françois Witz]
{\parbox{\textwidth}{\centering Adrien Schockaert$^{1,2,3}$\orcid{0009-0005-5717-0144}
        Hamid Laga$^4$\orcid{0000-0002-4758-7510} 
        Hazem Wannous$^1$\orcid{0000-0001-8475-4309}
        Vincent Magnier$^2$
        Guillaume Dufaye$^3$
        and Jean-françois Witz$^2$
        }
        \\
{\parbox{\textwidth}{\centering $^1$CERI SN, IMT Nord Europe, Villeneuve D'Ascq, 59650 (France)\
         $^2$ Univ. Lille, CNRS, Centrale Lille, UMR 9013 - LaMcube - Laboratoire de Mécanique, Multiphysique, Multiéchelle, F-59000 Lille, France \
         $^3$ Downs, 59670 Sainte-Marie-Cappel (France) \
         $^4$  School of Information Technology, Murdoch University (Australia)
       }
}
}

\begin{document}

\maketitle

\begin{abstract}
  The selection of an appropriate 3D representation is a fundamental design decision that dictates the efficiency, quality, and capabilities of modern computer vision and graphics pipelines for tasks such as 3D reconstruction, novel-view synthesis and rendering, shape and motion analysis, recognition, and generation. While traditional representations (\eg meshes, point clouds, and volumetric grids) remain standard outputs of 3D sensors (\eg LiDAR and 3D scanners) and are widely used in downstream applications (\eg editing and simulation), recent neural and primitive-based representations (\eg 3D Gaussian Splatting) offer compact and differentiable alternatives opening a wide range of opportunities in applications such as games, AR/VR, autonomous driving, robot navigation, and medical imaging, to name a few. The goal of this paper is to survey the main families of 3D representations from discrete explicit formats to continuous implicit fields based either on neural rendering or primitive splatting. For each type of representation, we present the general formulation and its variants, discuss its benefits and limitations, and highlight key applications. We conclude the paper by outlining the open challenges and potential directions for future research. Distinct from recent surveys that broadly cover 3D object and scene reconstruction, this paper provides a focused analysis on the evolution of 3D representations themselves. We specifically emphasize the paradigm shift toward implicit representations, offering a novel perspective on how these emerging formats fundamentally alter 3D/4D workflows.

\ccsdesc[500]{Computing methodologies~Shape representations}
\ccsdesc[500]{Computing methodologies~Shape modeling}
\ccsdesc[300]{Computing methodologies~Machine learning}

\printccsdesc   
\end{abstract}  
\section{Introduction}
\label{sec:introduction}

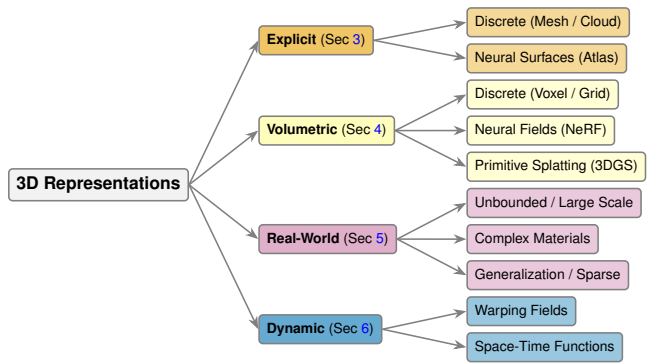
\begin{figure}[htbp]
  \centering
  \resizebox{\columnwidth}{!}{%
    \begin{forest}
      for tree={
          grow'=east,             
          draw=black!50,          
          thick,
          rounded corners=2pt,
          edge={draw=black!50, thick, ->, >=Stealth}, 
          l sep+=25pt,            
          s sep=3pt,              
          inner sep=4pt,          
          font=\sffamily\small,   
          anchor=west,            
          child anchor=west,
          parent anchor=east,
          calign=center,
          tier/.wrap pgfmath arg={tier#1}{level()}, 
      },
      root/.style={fill=gray!10, font=\sffamily\bfseries\large},
      explicit/.style={fill=cbOrange!60},      
      explicitsub/.style={fill=cbOrange!40},
      implicit/.style={fill=customYellow!80},     
      implicitsub/.style={fill=customYellow!50},
      realworld/.style={fill=customRose!60},  
      realworldsub/.style={fill=customRose!40},
      dynamic/.style={fill=cbTeal!60},    
      dynamicsub/.style={fill=cbTeal!40},
      [\textbf{3D Representations}, root
        [\textbf{Explicit} (Sec~\ref{sec: Explicit representations}), explicit
          [Discrete (Mesh / Cloud), explicitsub]
          [Neural Surfaces (Atlas), explicitsub]
        ]
        [\textbf{Volumetric} (Sec~\ref{sec:volumetric_representations}), implicit
          [Discrete (Voxel / Grid), implicitsub]
          [Neural Fields (NeRF), implicitsub]
          [Primitive Splatting (3DGS), implicitsub]
        ]
        [\textbf{Real-World} (Sec~\ref{sec:real_world}), realworld
          [Unbounded / Large Scale, realworldsub]
          [Complex Materials, realworldsub]
          [Generalization / Sparse, realworldsub]
        ]
        [\textbf{Dynamic} (Sec~\ref{sec:dynamic}), dynamic
          [Warping Fields, dynamicsub]
          [Space-Time Functions, dynamicsub]
        ]
      ]
    \end{forest}%
  }
  \caption{\textbf{Taxonomy and Organization.} A hierarchical overview of the survey structure. We categorize methods from Explicit Surfaces to Volumetric Fields, extending to Real-World and Dynamic applications.}
  \label{fig:taxonomy}
\end{figure}

Many computer vision and graphics tasks rely on an internal 3D representation of objects and scenes, including 3D reconstruction, novel-view synthesis and rendering, recognition and segmentation, shape and motion analysis, and content generation. The representation is a central design choice that determines what information is stored, how efficiently it can be learned from data, and how easily it can be queried, rendered, edited, or exported to downstream applications. Defining the right representation is thus crucial to the success of any 3D applications in fields such as robotics, autonomous driving, computer graphics, Virtual/Augmented Reality (VR/AR), sports analytics, medical diagnosis, injury rehabilitation, plant phenotyping, and beyond.

Traditionally, 3D information is acquired using active sensors (\eg LiDAR) or recovered using passive geometric methods (\eg multi-view stereo and Shape-from-X), typically producing explicit representations such as point clouds, meshes, or volumetric grids. In recent years, the rise of machine learning has shifted the focus toward formulating 3D tasks as learning problems. Therefore, rather than using discrete representations, many methods learn more modern learning-based representations that can be optimized directly from observations (\eg images, videos, depth, partial point clouds) and queried efficiently for rendering, editing, or downstream analysis. This shift makes the choice of representation central, as it dictates what information is encoded, how it is optimized from data, and which trade-offs are made between quality, efficiency, and applicability across tasks.

We emphasize that classical continuous geometric modeling representations widely used in computer graphics and CAD (e.g., spline and subdivision surfaces, moving least squares, and CSG) are outside the scope of this survey; accordingly, our comparisons are restricted to the learning-based reconstruction and neural rendering literature.

\begin{figure}[t]
    \centering
    \resizebox{\columnwidth}{!}{%
    \begin{tikzpicture}[
        font=\sffamily,
        >=Stealth,
        bubble/.style={circle, thick, fill opacity=0.9, align=center},
        axis/.style={->, thick, line width=1.2pt, color=black!80},
        lbl/.style={font=\bfseries\footnotesize, text=black!80, align=center},
        cSurface/.style={fill=customRose!70},      
        cVolumetric/.style={fill=cbTeal!70},  
        cHybrid/.style={fill=cbOrange!70},    
        sContinuous/.style={draw=black!80, line width=1.0pt},
        sDiscrete/.style={draw=black!80, line width=1.0pt, dash pattern=on 3pt off 2pt},
    ]

    \draw[axis] (0,0) -- (10.5,0) node[midway, below=8pt, font=\bfseries] {Downstream Compatibility \& Speed};
    \draw[axis] (0,0) -- (0,6.5) node[midway, above=10pt, rotate=90, font=\bfseries] {Representation Fidelity};

    \begin{scope}[shift={(0.5, 5.5)}]
        \node[anchor=west, font=\bfseries\scriptsize] at (0, 0.8) {\textsc{Class (Color)}};
        
        \draw[cSurface, draw=none] (0.2, 0.4) circle (0.12);
        \node[anchor=west, font=\scriptsize] at (0.4, 0.4) {Surface};
        
        \draw[cVolumetric, draw=none] (1.8, 0.4) circle (0.12);
        \node[anchor=west, font=\scriptsize] at (2.0, 0.4) {Volumetric};
        
        \draw[cHybrid, draw=none] (3.8, 0.4) circle (0.12);
        \node[anchor=west, font=\scriptsize] at (4.0, 0.4) {Hybrid};

        \node[anchor=west, font=\bfseries\scriptsize] at (6.0, 0.8) {\textsc{Domain (Line)}};
        
        \draw[fill=gray!20, sContinuous] (6.2, 0.4) circle (0.12);
        \node[anchor=west, font=\scriptsize] at (6.4, 0.4) {Continuous};

        \draw[fill=gray!20, sDiscrete] (8.2, 0.4) circle (0.12);
        \node[anchor=west, font=\scriptsize] at (8.4, 0.4) {Discrete};
    \end{scope}


    \node[bubble, cSurface, sDiscrete, minimum size=1cm] (points) at (7.5, 1.0) {};
    \node[lbl, above=1pt] at (points.north) {Points};

    \node[bubble, cSurface, sDiscrete, minimum size=1.0cm] (mesh) at (9, 1.5) {};
    \node[lbl, above=1pt] at (mesh.north) {Meshes};
    
    \node[bubble, cSurface, sContinuous, minimum size=0.5cm] (maps) at (5.5, 1.8) {};
    \node[lbl, above=1pt] at (maps.north) {Surface maps};

    \node[bubble, cVolumetric, sDiscrete, minimum size=1.5cm] (voxel) at (1.5, 1.1) {};
    \node[lbl, above=1pt] at (voxel.north) {Voxels};

    \node[bubble, cVolumetric, sContinuous, minimum size=0.5cm] (nerf) at (2.0, 4.6) {};
    \node[lbl, above=1pt] at (nerf.north) {Radiance\\field};

    \node[bubble, cVolumetric, sContinuous, minimum size=0.5cm] (surface) at (2.5, 3.3) {};
    \node[lbl, above=1pt] at (surface.north) {Distance\\field};

    \node[bubble, cHybrid, sContinuous, minimum size=1.0cm] (gs) at (7.5, 4.0) {};
    \node[lbl, above=1pt] at (gs.north) {Primitives\\splatting};

    \end{tikzpicture}%
    }
    \caption{\textbf{The Landscape of 3D Representations.} 
    We categorize methods by \textbf{Class} (Color) and \textbf{Domain Nature} (Line Style).
    \textbf{Solid lines} indicate Continuous neural fields (infinite resolution), while \textbf{dashed lines} indicate Discrete structures.
    Node size represents the memory footprint.}
    \label{fig:landscape}
\end{figure}
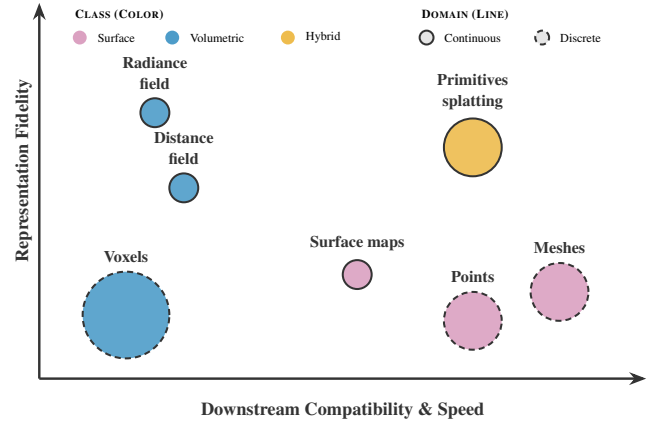

Existing surveys primarily review architectures and training mechanisms for specific tasks and supervision regimes~\cite{han2019image, guo2021deep, xie2022neural, tian2023recovering, Sulzer2025surface}. In this survey, we focus specifically on the representation choices made in learning-based pipelines and how they affect optimization, differentiability, rendering, memory, and downstream compatibility. We visually map this landscape in Fig.~\ref{fig:landscape}, categorizing representations based on the trade-offs they offer between fidelity and deployability. Traditional discrete formats (meshes, point clouds) continue to dominate industrial workflows and sensing pipelines due to their interpretability and direct compatibility with physics engines. However, they can be memory-intensive or difficult to integrate into learning and differentiable rendering pipelines. Recent alternatives, including continuous neural fields and primitive-based representations such as Gaussian splatting, aim to improve compactness and enable efficient optimization and rendering while introducing new trade-offs.

This paper presents a unified taxonomy of learning-based 3D representations and uses it to clarify the design space and practical trade-offs (accuracy, efficiency, differentiability, editability, and downstream compatibility). We begin by formulating the problem and introducing the taxonomy (Section~\ref{sec:problem_statement}). We then review explicit surface representations (Section~\ref{sec: Explicit representations}) and implicit volumetric representations (Section~\ref{sec:volumetric_representations}), before addressing extensions required in real-world settings (Section~\ref{sec:real_world}) and for dynamic scenes (Section~\ref{sec:dynamic}). Finally, we survey representative applications (Section~\ref{sec:applications}) and conclude with open challenges and future directions (Section~\ref{sec:future_research}). To maintain a focused scope on learning-based reconstruction, we exclude classical continuous modeling techniques (e.g., B-Splines, CSG) except where they intersect with neural pipelines.

\section{Taxonomy and problem statement}
\label{sec:problem_statement}

\begin{figure*}[t]
\centering
    \includegraphics[width=\textwidth]{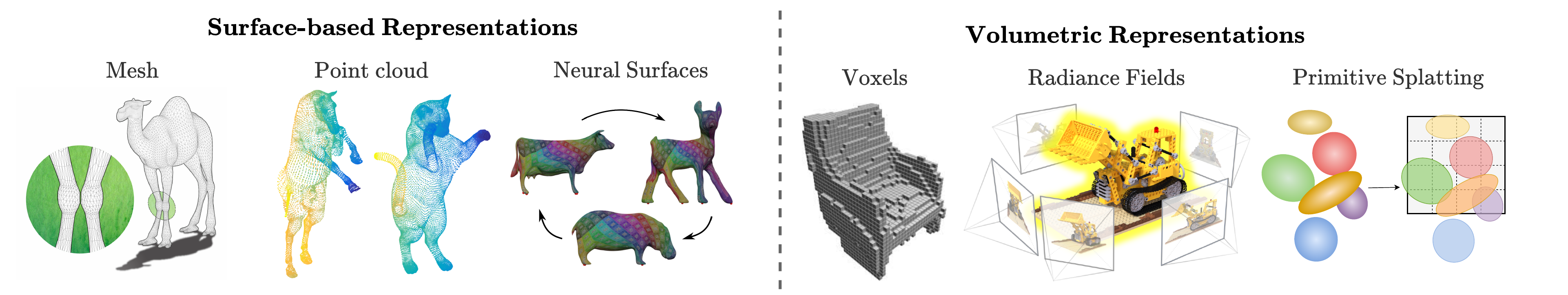}
    \caption{\textbf{Visual Overview of 3D Representations.} 
    Spectrum displaying the different representations discussed in this survey from Surface-based (left), discussed in Sec~\ref{sec: Explicit representations}, to recent Volumetric (right) representations, discussed in Sec~\ref{sec:volumetric_representations}.}
    \label{fig:spectrum}
\end{figure*}

Existing methods for 3D shape, motion, and appearance reconstruction addressed various aspects of the problem, from the representation and network architecture to the training mechanism and mode of supervision. This paper focuses on the representation and its importance for the downstream tasks. Fig.~\ref{fig:taxonomy} provides an overview of the structure of the paper and a taxonomy of the state-of-the-art.

In this survey, we define 3D representations as queryable models of the form:
\begin{equation}
\begin{split}
    \neuralsurfacefunc_\params &: \domain \rightarrow \mathcal{S}, \\
    \shapeproperties &= \neuralsurfacefunc_\params\big(\gamma(\mathbf{q})\big), 
\end{split}
\label{eq:3d_representation}
\end{equation}

\noi \noindent Given a query $\inputparams$ (optionally processed by an encoding $\enc$), the function returns a set of properties $\shapeproperties$. The domain $\mathcal{D}$ is representation-dependent: it may span continuous spaces (e.g., 3D coordinates $\domainpoint$, viewing directions $\viewdir$, time $t$) or discrete sets (e.g., voxel, pixel, or primitive indices). Similarly, the output space $\mathcal{S}$ encompasses diverse attributes, ranging from geometry (e.g., occupancy, signed distance, position) to appearance (e.g., radiance, spherical harmonics, texture). 
This unified formulation bridges the gap between explicit/discrete representations—where $\params$ consists of stored coefficients (vertices, voxel densities) accessed via lookup and interpolation—and implicit/neural ones, where $\params$ parameterizes a function (MLP) to regress properties continuously. The fundamental distinction lies in whether the mapping is structurally explicit or implicit, and whether its domain is discrete or continuous. Figure~\ref{fig:spectrum} provides a qualitative comparison of the representation covered by this formulation.

\textbf{Explicit representations (Section~\ref{sec: Explicit representations})} define the mapping $\neuralsurfacefunc$ using an explicit function of the form $\neuralsurfacefunc: \domain  \to \rthree \times \appearancespace$. When the domain $\domain$ is discretized, we obtain representations such as polygonal meshes and point clouds (Section~\ref{sec:discrete_explicit_representations}). Neural surfaces (Section~\ref{sec:neural_surfaces}), on the other hand, operate directly on the continuous domain, without requiring upfront discretization of the domain  $\domain$.  Continuous explicit representations are compact and resolution-agnostic; thus, in theory, they can represent 3D objects at any arbitrary level of detail. However, they are typically limited to objects of fixed topology, usually of low genus.

\textbf{Implicit representations} (Section~\ref{sec:volumetric_representations}) such as volumetric occupancy or signed distance grids can represent objects and scenes of arbitrary topologies. Their advantage lies in being regular structures and thus can be processed using convolutional operations. In their discrete form, implicit representations are computationally expensive due to high memory requirements. Their full potential, however, emerges when using continuous queryable models, such as neural fields.
\textbf{Unbounded scenes representations} (Section~\ref{sec:real_world}) address the spatial limitations of 3D representations. While conventional methods are often restricted to fixed-size, bounded domains, real-world environments — such as city-scale reconstructions or open landscapes — require representations that can adapt to varying spatial extents efficiently, without recurring extensive memory or computational costs.

\textbf{Dynamic representations} (Section~\ref{sec:dynamic}) extend 3D representations to capture time-varying geometry and appearance. These methods not only represent static 3D objects but also encode their motion and deformation over time, which is crucial for modeling real-world scenes.

\section{Explicit Representations}
\label{sec: Explicit representations}

Explicit representations describe 3D geometry by directly parameterizing surface locations in the spatial domain. Formally, they define the surface as an explicit mapping function $\neuralsurfacefunc$ that associates each point $\domainpoint \in \domain$ with its corresponding 3D coordinate $\point \in \geometryspace = \rthree$. Depending on the nature of the domain $\domain$, explicit representations can be broadly classified into two categories: \textbf{(1)} discrete representations, which approximate geometry using finite sets of samples such as polygonal meshes or point clouds (\Cref{sec:discrete_explicit_representations}), and \textbf{(2)} continuous representations, which rely on learnable functions to define surfaces in a smooth and differentiable manner (\Cref{sec:neural_surfaces}). This section discusses both classes, their underlying principles, and their respective role in modern 3D modeling pipelines. Table~\ref{tab:explicit_compare} provides a structural comparison of these representations, highlighting their fundamental differences in terms of continuity, connectivity, and topological flexibility.

\begin{table*}[t]
\centering
\scriptsize
\renewcommand{\arraystretch}{1.6}
\setlength{\tabcolsep}{6pt}
\rowcolors{2}{gray!10}{white}

\begin{tabularx}{\textwidth}{l c c c X X l}
\toprule
\textbf{Representation} & 
\multicolumn{3}{c}{\textbf{Properties}} & 
\textbf{Strengths \& Benefits} & 
\textbf{Weaknesses \& Limitations} & 
\textbf{Key References} \\
\cmidrule(lr){2-4}
& \tiny{Continuity} & \tiny{Connectivity} & \tiny{Topology} & & & \\
\midrule

\textbf{Polygonal Meshes} & 
\high & \high & \low & 
Standard for rendering; Explicit control over vertex density; Efficient rasterization. & 
Irregular graph structure complicates DL (requires GNNs); Non-differentiable topology changes. & 
MeshCNN~\cite{hanocka_meshcnn_2019}, Pixel2Mesh~\cite{wang2018pixel2mesh} \\

\textbf{Point Clouds} & 
\midlow & \low & \high & 
Native sensor output; Trivial to resize/manipulate; No connectivity overhead. & 
No surface definition (holes); Requires radius tuning for rendering; Hard to model fine flat surfaces. & 
PointNet~\cite{qi2017pointnet}, Pulsar~\cite{lassner2021pulsar} \\

\textbf{Neural Surfaces} & 
\midlow & \low & \low & 
\textbf{Infinite resolution;} Differentiable w.r.t parameters; Intrinsically regular (UV space) for CNNs. & 
Restricted topology (cannot represent holes/complex genus without multiple charts). & 
AtlasNet~\cite{groueix2018papier}, FoldingNet~\cite{yang2018foldingnet} \\

\bottomrule
\end{tabularx}

\caption{\textbf{Structural Comparison of Explicit Representations.} 
This table classifies representations based on their mathematical properties:
\textbf{Continuity} (Is the surface defined between samples?), 
\textbf{Connectivity} (How are samples related?), and 
\textbf{Topology} (Can it represent arbitrary genus/holes?). It also gives strengths and weaknesses of each category with key references.\\
\emph{Legend:} \quad \high~High (Native/Excellent) \qquad \midlow~Medium (Possible/Limited) \qquad \low~Low (Poor/Unsuited).
\label{tab:explicit_compare}}
\end{table*}

\subsection{Discrete Explicit Representations}
\label{sec:discrete_explicit_representations}
Discrete explicit representations approximate 3D objects and scenes by discretizing the domain $\domain$ and encoding surfaces as polygonal meshes or point clouds. Despite recent advances in continuous representations, discrete formats remain indispensable in computer graphics. This is due to their deep-rooted integration into real-world workflows: they are the standard in industry 3D software, animation pipelines, gaming engines, visual effects production, and physics simulation. 
Discrete representations are also the native data format of 3D sensors, including LiDAR, RGB-D cameras, and structured-light scanners.
While discrete representations are mathematically explicit and efficient for rendering, integrating them into neural networks presents fundamental challenges. Unlike 2D images or 3D voxel grids, meshes and point clouds lack a regular, grid-aligned structure, rendering standard convolution and pooling operations, which are crucial in many neural networks, inapplicable. Methods must instead account for the specific irregularities of the data, requiring permutation invariance and the ability to process highly variable, non-uniform, and unordered structures. Moreover, raw 3D Cartesian coordinates are highly sensitive to transformations (\ie rotation and translation); a network trained purely on raw coordinates will likely fail if the object is simply rotated or translated. Consequently, adapting these formats for deep learning requires specialized backbones capable of extracting mathematically invariant geometric features directly from unstructured data.

\subsubsection{Mesh-based representations}

\begin{figure}[htbp]
\centering
    \includegraphics[trim={0cm 0cm 9cm 0cm}, clip, width=\columnwidth]{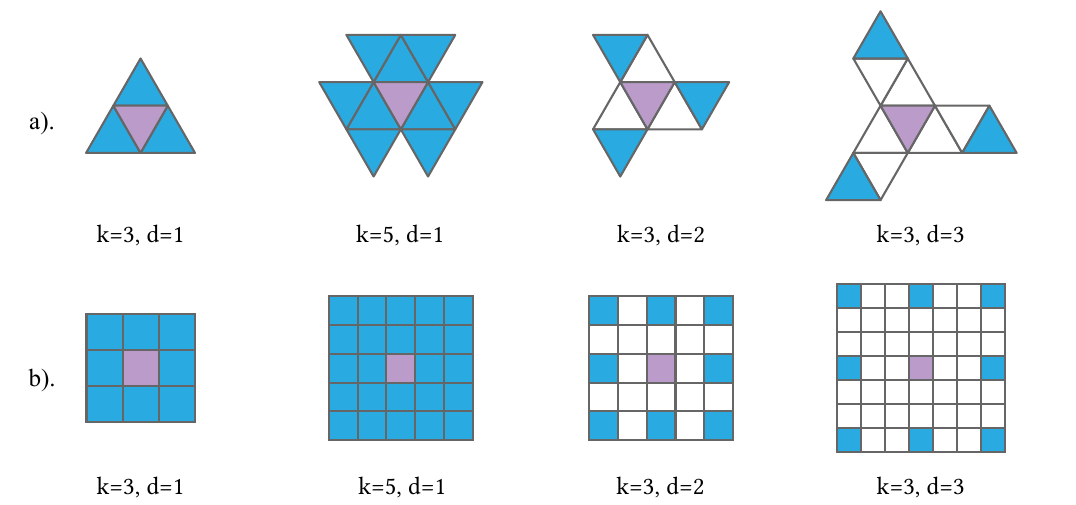}
    \caption{Mesh convolution kernel pattern. (a) Mesh convolution kernels with different kernel sizes and dilation. (b) Corresponding 2D convolution kernels. figure borrowed from~\cite{hu2022subdivision}.} 
    \label{fig:MeshConvolution}
\end{figure}

In the context of our unified formulation (Eq.~\ref{eq:3d_representation}), polygonal meshes act as a piecewise linear approximation of the function $\neuralsurfacefunc_\params$. Here, the domain $\domain$ is the discrete set of faces, the query $\mathbf{q}$ is defined by a face index and its local coordinates, and the output space $\mathcal{S}$ is the corresponding 3D spatial coordinate $\point \in \mathbb{R}^3$. The representation parameters $\params$ are explicitly stored as the 3D vertex positions and connectivity graph, while the encoding $\gamma$ is simply the identity function. Polygonal meshes provide explicit surface topology through vertices, edges, and faces, utilizing non-uniform polygons to represent both large flat areas and intricate high-frequency details. Any point on the surface can be defined using its barycentric coordinates relative to the vertices of the polygonal face to which it belongs to. While meshes facilitate efficient storage, manipulation, and rendering, their non-uniformity creates irregular graph structures that inhibit conventional grid-based operations used in deep learning. A key challenge is then to design neural architectures that can effectively process and learn from such irregular data structures. Methods aiming to answer this challenge can be broadly classified based on the fundamental geometric primitive they treat as their primary data structure:

\vspace{6pt}
\noi\textbf{Vertex-based Methods.} By defining processes primarily on mesh vertices, several methods have defined mesh-specific architectures. Some methods~\cite{masci_geodesic_2015, boscaini_learning_2016, tatarchenko2017octree, poulenard2018geodesicNeuralNet} define a local parametrization of the mesh by mapping geodesic patches or neighborhoods around sampled vertices into regular 2D domains (\ie tangent planes) where standard 2D Convolutional Neural Networks (CNNs) can be applied to learn the local geometry. Others~\cite{wang2018pixel2mesh,chen2021meshconv,dai2019scan2mesh} treat the mesh as a standard graph, with vertices as nodes connected by edges, enabling the use of Graph Neural Networks (GNNs) to aggregate features based on vertex connectivity. However, GNNs often ignore the underlying geometry and require multiple layers to aggregate global features. To avoid these limitations, methods can leverage continuous, mathematically grounded differential operators (\ie Hodge Laplacian~\cite{smirnov_hodgenet_2021}, Laplace-Beltrami~\cite{sharp2022diffusionnet}), thereby efficiently supporting global spatial support without deep layer stacking.

\vspace{6pt}
\noi\textbf{Edge-based Methods.} These methods leverage the topological guarantee that in a closed 2-manifold mesh, every edge is incident to exactly two faces and adjacent to four edges. This property is exploited by Hanocka \etal~\cite{hanocka_meshcnn_2019} to define an ordering invariant convolution. Alternatively, features can be extracted by taking random walks~\cite{lahav2020meshwalker} along the edges of the mesh. Milano \etal~\cite{milano_primal-dual_2020} adapt edge-based processing to a graph by constructing both a primal graph (representing faces) and a dual graph (representing edges). They perform dynamic feature aggregation using an attention mechanism and implement pooling by contracting edges in the primal graph, which effectively clusters the faces of the mesh in a task-specific manner.

\vspace{6pt}
\noi\textbf{Face-based Methods.} Face-based methods~\cite{feng2019meshnet,10.1145/3474085.3475468,hu2022subdivision} focus on the triangular polygons themselves. The Pioneer MeshNet~\cite{feng2019meshnet} established the face-based paradigm by designing a convolution block that directly operates on faces. To understand local geometry, it introduced a Face Kernel Correlation (FKC) descriptor that measures the geometric affinities between the face normals in a local neighborhood and a set of reference kernels. However, the method is very shallow; it completely lacks pooling layers to downsample features, and because a normal's value is flat across an entire face, the FKC descriptor struggles to thoroughly capture complex local structures. This was enhanced in MeshNet++~\cite{10.1145/3474085.3475468}, which redesigned the face-centric approach for a deeper and more expressive architecture. They define a decoupled feature representation through a spatial and a structural descriptor, producing features that are then concatenated into a comprehensive representation. Moreover, they define a surface correlation block that samples multiple points directly from the surface of neighboring faces using barycentric coordinates. This helps to effectively learn multi-scale local structures while disentangling "how" a shape looks from "where" it is located. Alternative methods perform a remeshing step to create a regular, hierarchical structure. Hu \etal proposed SubdivNet~\cite{hu2022subdivision}, a method in which they use self-parametrization to perform remeshing to enforce subdivision sequence connectivity. This connectivity guarantees that every face is surrounded by exactly three adjacent faces, effectively creating a uniform mesh pyramid, analogous to a 2D image pyramid (see Fig~\ref{fig:MeshConvolution}). This structural regularity allows to define a highly flexible mesh convolution operator that supports variable kernel sizes, strides, and dilations. Furthermore, pooling and upsampling are cleanly defined as uniform 4-to-1 face merging and 1-to-4 face splitting operations.

\vspace{6pt}
\noi\textbf{Sequence-based Methods.}
Driven by advancements in natural language processing, the most recent architectures serialize the 3D mesh into a 1D sequence, utilizing large transformers or State-Space Models (SSMs) to capture global context. MeshGPT~\cite{siddiqui2024meshgpt} treats the mesh as a sequence of triangles to perform autoregressive prediction of the next triangle token, allowing the model to generate clean, sharp, and cohesive 3D meshes. Alternatively, methods leverage SSMs~\cite{zhang2025mesh,yoshiyasu2025meshmamba}, processing the token sequence with Mamba blocks and feature diffusion, they capture intricate spatial relationships and highly content-aware global context with linear computational complexity.

\subsubsection{Point-based Representations}\label{sec:point_rep}

The use of points as a fundamental primitive for 3D representation has a long history in computer graphics~\cite{csuri1979towards,levoy1985use}. Mathematically, a point cloud is an unordered set of 3D coordinates $\mathcal{P} = \{\mathbf{p}_i \in \mathbb{R}^3 \mid i=1, \dots, N\}$, often augmented with attributes such as normals, colors, or sensor intensities. Mapping this to Eq.~\ref{eq:3d_representation}, the domain $\domain$ is a discrete index set $\{1, \dots, N\}$, and the query $\mathbf{q}$ is a specific point index. The mapping function $\neuralsurfacefunc_\params$ behaves as a discrete lookup table where the parameters $\params$ store the 3D geometry and attributes of each point, yielding the output space $\mathcal{S} \subset \mathbb{R}^3 \times \appearancespace$.

\begin{figure*}[htbp]
\centering
    \includegraphics[width=0.9\textwidth]{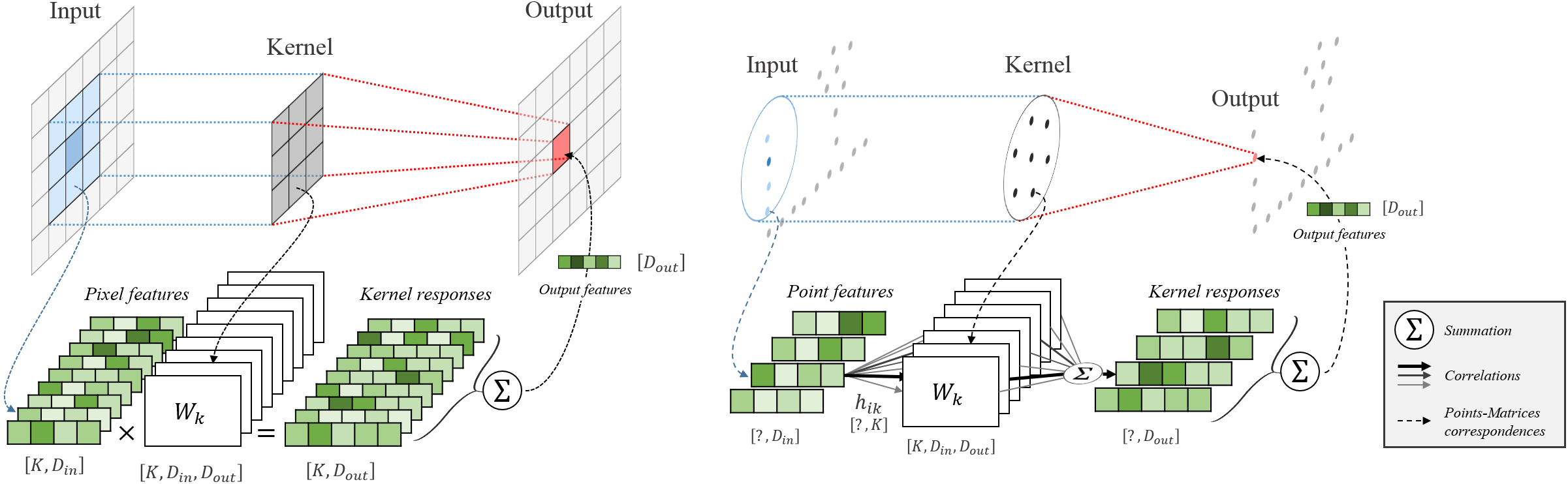}
    \caption{Comparison between classical convolution (left) and point-based convolution (right). Image borrowed from~\cite{Thomas_2019_ICCV}}. 
    \label{fig:PointConvolution}
\end{figure*}

\vspace{6pt}
\noi\textbf{Learning on Point Sets.}
A primary challenge in processing point clouds is achieving invariance to geometric transformations, particularly permutation (changing the order of points should not change the output). PointNet~\cite{qi2017pointnet} introduced a permutation-invariant architecture using shared Multi-Layer Perceptrons (MLPs) followed by a global max-pooling operation. While this effectively learns global descriptors, it fails to capture local geometric context. Subsequent works~\cite{qi2017pointnet++,li2018pointcnn,Thomas_2019_ICCV,Wu_2019_CVPR} addressed this limitation by introducing hierarchical learning. They partition the point set into nested local neighborhoods, applying MLPs recursively to abstract features at increasing scales to generalize feature extraction. Alternatively, KPConv~\cite{Thomas_2019_ICCV} operates in continuous Euclidean space, defining convolution weights at specific kernel points influenced by a linear correlation function (see Fig~\ref{fig:PointConvolution}). Extracting these hierarchical features requires the definition of a downsampling process. While this is typically done via Farthest Point Sampling (FPS), methods enhance the process, relying on grid subsampling~\cite{Thomas_2019_ICCV} or partition-based grid pooling~\cite{wu2022point}. Doing so, they ensure a sampling that is robust to the variation in point density that can come from the sensor.
Beyond MLPs, recent approaches leverage the advances in deep learning and advance other architectures for point cloud processing. Some approaches~\cite{wang2019dynamic} model points as a graph, dynamically updating neighbors in feature space. Other leverage modern transformers~\cite{zhao2021point,wu2022point,wu2024point} or state-space models~\cite{liang2024pointmamba,zha2025pma,li2026dapointmamba}. They deliberately break permutation invariance, structuring data via space-filling curves~\cite{wang2023octformer} or using serialized encoding~\cite{wu2024point,liang2024pointmamba}. As standard sequence positional embeddings fail on sparse point clouds, architectures require custom positional awareness obtained through a sparse convolutional layer~\cite{wu2024point} or direct spatial coordinate mapping via a linear layer~\cite{liang2024pointmamba}.

\vspace{6pt}
\noi\textbf{Surface Reconstruction from Points.}
Extracting a watertight manifold mesh from raw, unstructured point clouds is a fundamental problem generally tackled through two paradigms. Optimization-based methods fit a surface directly to a single point cloud without prior training. Classical approaches like Poisson Surface Reconstruction~\cite{kazhdan2013screened} remain the gold standard for clean, dense data. However, they struggle with sparse or heavily corrupted inputs. To handle such data, learning-based methods leverage geometric priors acquired from large 3D datasets. By encoding the unstructured point cloud into intermediate local or hierarchical structures, such as voxel grids~\cite{peng2020convolutional, boulch_poco_2022}, octrees~\cite{ummenhofer_adaptive_2021}, or local patches~\cite{siddiqui2021retrieval, erler2024ppsurf}, these networks can robustly synthesize high-frequency geometric details while maintaining global topological coherence.

\vspace{6pt}
\noi\textbf{Differentiable Rendering and Neural Points.}
For synthesis tasks, raw points lack connectivity, leading to holes and aliasing. To mitigate this, classical graphics introduced surfels (surface elements). In learning-based reconstruction, this concept evolved into \textit{differentiable point representations}. Methods like Neural Points~\cite{aliev2020neural, lassner2021pulsar, ruckert2022adop} augment each point with a high-dimensional feature vector projected and rasterized into screen space. These methods serve as the direct foundation for breakthroughs in primitive-based splatting~\cite{Kerbl_2023_3dGS} discussed in Sec~\ref{sec:3dgs}.

\subsection{Neural Surfaces}
\label{sec:neural_surfaces}

Due to their discrete nature, representing fine geometric details using polygonal meshes and point clouds requires high-resolution sampling of the 3D space, which can be expensive in terms of memory requirements. An alternative is to leverage the classical idea from differential geometry stating that surfaces are 2D manifolds embedded in $\mathbb{R}^3$ and therefore admit locally planar parameterizations. Concretely, they are then represented as a collection of \emph{charts} (surface patches), each parameterized by a continuous mapping from a 2D domain to 3D space. Together, these charts form an \emph{atlas} that covers the entire surface. Following this idea, \emph{Atlas-based representations} learn a collection if charts $\{\neuralsurfacefunc_{\params_i}: \domain_i \subset \mathbb{R}^2 \rightarrow \mathbb{R}^3, i=1, \dots, N\}$ such that their union represent the surface. Seminal work from Groueix \etal~\cite{groueix2018papier} introduced the idea of learning such parameterizations using neural networks. Specifically, they parametrize each chart of the atlas with a neural network. Under our framework (Eq.~\ref{eq:3d_representation}), the domain $\domain \subset \mathbb{R}^2$ is a continuous 2D manifold (e.g., a unit square or sphere). The query $\mathbf{q}$ is a 2D coordinate $\domainpoint$, and the parameters $\params$ are the learnable weights of the Multilayer Perceptron (MLP). The function $\neuralsurfacefunc_\params$ maps these 2D queries directly to the 3D shape, defining the output space $\mathcal{S} \subset \mathbb{R}^3$. Their model, referred to as \textit{AtlasNet}, can represent surfaces of arbitrary topologies in a continuous and differentiable manner. However, each chart is defined independently, which can lead to inconsistencies and distortions across overlapping regions. Subsequent works~\cite{yuan2018pcn,williams2019deep,zhao20193d,bednarik2020shape,badki2020meshlet,Ma_2021_CVPR,lin_pctma-net_2021} have proposed various strategies to address these issues, such as enforcing consistency across overlapping charts~\cite{bednarik2020shape}, using local shape priors~\cite{badki2020meshlet}, or using more sophisticated architectures to model the relationships between charts~\cite{lin_pctma-net_2021}. The chart-specific MLPs  can have shared weights~\cite{yuan2018pcn,Ma_2021_CVPR,ma2021power}  or be completely independent~\cite{sinha2016deep,groueix2018papier,zhao20193d,williams2019deep,bednarik2020shape}, the latter offering a better fit to 3D shapes. However, the number of network parameters, and thus the memory cost, increases linearly with the number of patches, which is a big limit for real-life applications.

Orthogonally, surfaces can be represented using a single chart that covers the entire surface~\cite{yang2018foldingnet,pang2021tearingnet,zhang2021ners}. For instance, FoldingNet~\cite{yang2018foldingnet} starts from a 2D grid and learns to deform it into the target 3D shape using a neural network. TearingNet~\cite{pang2021tearingnet} extends this idea by learning a tearing network that interacts with the folding network by tearing the representation. Essentially, the tearing network parametrizes the surface defined by the folding network according to a learned topology.

Most of Neural Surfaces generally define the domain mapped to $\mathbb{R}^3$ as a square or a disk. However, this choice limits the representation to genus-0 surfaces. To overcome this limitation, some methods assume spherical shape topologies and use more advanced domains, such as a torus~\cite{ben2018multi} or sphere~\cite{zhang2021ners,cheng2022diffeomorphic,williamson2025neural}. Others~\cite{groueix20183d,Ma_2021_CVPR} specialized in representing human shapes by learning to deform a template to a target human shape.

While being an efficient continuous representation of surfaces, neural surfaces are limited to representing shapes. In the absence of native geometric operators for neural representations, the common workaround for applying geometry processing, analysis, or editing is to work on the maps or by discretizing these representations into an explicit mesh (\eg using Marching cubes) and then relying on traditional mesh-based operators. Recent methods have started to explore methods working directly on the neural surface representation~\cite{morreale2021neural,morreale2022neural,morreale2024neural} for composition~\cite{morreale2021neural,morreale2022neural}, semantic matching~\cite{morreale2024neural}, and discretization-free geometry processing~\cite{novello_exploring_2022,yang2021geometry,novello_neural_2023,williamson2025neural}.

\section{Volumetric Representations}
\label{sec:volumetric_representations}

Volumetric representations define the domain $\domain$ of Eqn.~\eqref{eq:3d_representation} as a subset of $ \rthree$, \eg $[0, 1]^3$. The shape properties at each point $\domainpoint \in \domain$ are then defined implicitly, in the form of:

$\bullet{}$ \textit{Occupancy}, which encodes the probability of $\domainpoint \in \domain$ being inside or outside the 3D shape~\cite{huang2018deepvolumetric,chen2019learning,mescheder2019occupancy}. Occupancy can also be binary, \ie a point $\domainpoint$ is set to one if it is inside the object of interest, and to zero otherwise. 

$\bullet{}$ \textit{Distance functions}, which assign to each point $\domainpoint \in \domain$ its distance to the closest point on the shape's surface. Usually, methods employ a Signed Distance Function (SDF)~\cite{park2019deepsdf, wang2021neus, wang2023neus2, li2023neuralangelo} where the sign indicates whether the point is inside or outside the object.

Theoretically, this continuous representation allows the extraction of geometry at any resolution. However, with its clear definition of the inside and the outside of an object, the SDF is restricted to watertight surfaces. In the case of open surfaces, it is preferable to learn a Truncated Signed Distance Function (TSDF)~\cite{curless1996volumetric, sun2021neucon} or to use the Unsigned Distance Function (UDF)~\cite{liu2023neudf, long2023neuraludf, zhou2024cappami, Fainstein_2024_CVPR}.

Note that one can also store at each point $\domainpoint$, its appearance in the form of a texture value (RGB) or radiance (RGB and volume density), which are view-dependent. In such a case, the representation is referred to as a radiance field.

Volumetric representations have traditionally relied on discretized 3D voxel grids, where scene properties are stored at the center of each grid cell or voxel~\cite{han2019image}. While such grids are well-suited for processing and inference using convolutional operations, their computational efficiency decreases at high resolutions: representing surface details and fine structures will necessitate excessively large grids. To mitigate this, hierarchical space partitioning methods, such as Octrees~\cite{riegler2017octnet,wang2017cnn,tatarchenko2017octree,li2017grass}, exploit spatial sparsity to reduce memory overhead. Nevertheless, these approaches inherit the fundamental limitations of discrete representations, including resolution-dependent artifacts and high memory requirements. Moreover, reconstructing high-fidelity scenes from voxel grids demands heavy convolutional decoders, further constraining scalability.                             
Recent continuous implicit representations bypass these challenges by modeling scenes as continuous functions that map spatial coordinates directly to scene properties; see Eqn~\ref{eq:3d_representation}. These representations are resolution-agnostic, memory-efficient, and can encode intricate geometries with high accuracy. Their continuous nature also enables an end-to-end differentiation, facilitating seamless integration with modern architectures for joint optimization of geometry and appearance~\cite{mildenhall2020nerf, Kerbl_2023_3dGS}.

This section analyzes the evolution of implicit representations in 3D reconstruction. We begin with classical approaches based on Radial Basis Functions (Section~\ref{sec:radial_basis_function}), which lay the foundation for continuous scene approximation. Next, we explore neural implicit functions (Section~\ref{sec:neural_implicit_function}), where MLPs are used as universal approximators of continuous functions. Section~\ref{sec:neural_radiance_fields} reviews neural radiance fields, which extend this paradigm to view-dependent volumetric rendering, enabling the joint representation of geometry and appearance. Finally, Section~\ref{sec:3dgs} discusses 3D Gaussian Splatting, a recent breakthrough that optimizes 3D Gaussian within a differentiable rendering pipeline, achieving state-of-the-art trade-offs in fidelity and rendering speed.

\subsection{Radial Basis Functions (RBF)}
\label{sec:radial_basis_function}
Traditionally, continuous implicit functions have been parameterized using Radial Basis Functions (RBFs)~\cite{carr2001reconstruction,kojekine2004surface,ohtake2005multi}. An RBF is a function of the form:
\begin{equation}
     \begin{split}
        \neuralsurfacefunc: &  \rthree \to \real,  \text{ such that } \neuralsurfacefunc(\point) = \polynomial(\point) + \sum_{i=1}^\npoints \alpha_i\basicfunction_i(\| \point - \point_i \|).
    \end{split}
\end{equation}

\noi Here,  $\polynomial$ is a polynomial of low degree,  $\basicfunction_i$ are  real-valued functions on $[0, \infty)$, and  $\alpha_i$ are blending weights.  $\basicfunction_i$  are also referred to as basis functions and can be of infinite~\cite{carr2001reconstruction} or compact~\cite{kojekine2004surface} support. The points $\point_i$ are the centers of the RBFs.
Popular choices of the basis function $\basicfunction$ include (here $r = \| \point - \point_i \|$):
\begin{itemize}
    \item The thin-plate spline $\basicfunction(r) = r^2\log(r)$, which is suitable for fitting smooth functions of two variables,
    \item The Gaussian function $\basicfunction(r)  = \exp(-\frac{r^2}{2\sigma^2})$, which is suitable for neural networks. Gaussian can be isotropic (the way we defined it) or anisotropic with a covariance matrix~\cite{bouzidi_gnf_2025}.
    \item The multiquadratic $\basicfunction(r) = \sqrt{r^2 + c^2}$,  which is  particularly suitable for fitting topological data, 

    \item Biharmonic $\basicfunction(r) = r$ and triharmonic $\basicfunction(r) = r^3 $ splines, which are suitable for fitting functions of three variables.
\end{itemize}

\noi Commonly used basic functions in 3D surface reconstruction include the biharmonic function, the thin-plate spline, and the  Gaussian. The latter leads to RBF networks, which can be implemented as an MLP with an input layer, one hidden layer with a Gaussian as the activation function, and an output layer. Such RBF networks have been previously used for 3D object and light source representation~\cite{piperakis20013d} as well as for 3D object representation and reconstruction from dense point clouds~\cite{carr2001reconstruction,kojekine2004surface,ohtake2005multi}. In particular, Kojekine \etal~\cite{kojekine2004surface} use RBFs with compact support, significantly reducing the computation time both at training, since it results in a sparse matrix, and at runtime.

One of the main challenges when using RBFs is the choice of the basis function, as different functions have different properties and are suited for representing different geometric features. The challenge lies in the fact that a 3D object can have different types of geometric features, \eg smooth regions, corners, edges, and ridges. In that case, a single RBF type cannot model all these features.  To address this issue, Ohtake \etal~\cite{ohtake2005multi} leveraged the Partition of Unity (PoU) framework. Their method uses an octree structure to partition the input 3D shape, approximates each cell using a separate local shape function, and uses blending weights to smoothly combine them. A key feature of this approach is that it does not use a single pre-defined basis function. Instead,  at each cell of the octree, it selects, from a dictionary of basis functions, the one that best approximates the local geometry.

\subsection{Neural Implicit Representations}
\begin{figure}[htbp]
\centering
    \includegraphics[width=\columnwidth]{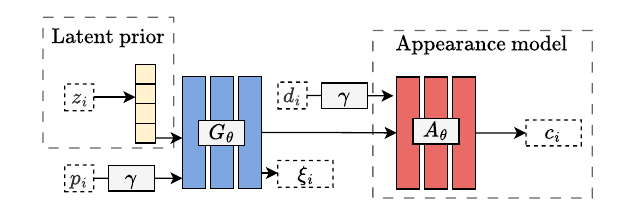}
    \caption{Architecture of a Neural Implicit Representation. The figure shows how neural implicit representations are articulated. A geometry network $G_\theta$ takes a spatial coordinate $\point_i$ and optionally a latent shape prior $\mathbf{z}_i$~\cite{park2019deepsdf} as input to predict a surface property $\shapeproperties_i$ (e.g., Signed Distance or Density). Additionally, an appearance network $A_\theta$ can utilize features extracted from $G_\theta$, along with the viewing direction $\mathbf{v}_i$, to regress the final color $\mathbf{c}_i$}. 
    \label{fig:implicit_architecture}
\end{figure}

\label{sec:neural_implicit_function}

Neural implicit functions, or neural fields, can be seen as a generalization of RBF networks to Multi-layer Perceptrons (MLPs), \ie feedforward networks with an arbitrary number of layers and without restricting the activation function to be a radial basis. The use of MLPs to implicitly represent 3D scenes has been explored previously.  For instance, in $2001$, Piperakis \etal~\cite{piperakis2001affine,piperakis20013d} demonstrated that both the 3D geometry of a scene and its light source could be approximated using an MLP. However, the performance, in terms of accuracy and computation time, was constrained by the limited computational power and training data available at that time.

In recent years, there has been a resurgence of these methods, where the 3D geometry of an object or scene is encoded as either a Signed Distance Function (SDF) or occupancy values~\cite{chen2019learning,mescheder2019occupancy}, and the neural function is parameterized using MLPs. Early methods, such as those proposed by Piperakis \etal~\cite{piperakis2001affine,piperakis20013d}, required training a separate network for each object or scene. To enable shape-agnostic representations, Park \etal~\cite{park2019deepsdf} represent the implicit field as a function of 3D points $\point \in \rthree$  and latent shape codes $\latentcode \in  \latentspace$ that contain all the information about the shape of the object. This formulation, known as DeepSDF, allows the representation of multiple SDFs with a single neural network; see Fig.~\ref{fig:implicit_architecture}. 

A key practical challenge in coordinate-based neural implicit functions is representing high-frequency variation: standard ReLU MLPs tend to exhibit a spectral bias toward low-frequency signals. Two widely used and largely orthogonal remedies are (i) \emph{Fourier feature} mappings that lift the input coordinates into a higher-dimensional sinusoidal space~\cite{tancik2020fourier}, and (ii) \emph{periodic-activation} networks such as SIREN~\cite{sitzmann2020implicit}, which use sinusoidal activation functions to better model fine detail without relying on an explicit input feature mapping.

\subsection{Neural Radiance Fields}
\label{sec:neural_radiance_fields}

Neural Radiance Fields (NeRF), introduced by Mildenhall \etal~\cite{mildenhall2020nerf},  extends early neural implicit representations~\cite{park2019deepsdf} to jointly model geometry and view-dependent appearance. In terms of our unified framework (Eq.~\ref{eq:3d_representation}), NeRF defines the query $\mathbf{q} = (\point, \viewdir)$ as a 5D input combining a 3D position $\point \in \domain = \rthree$ and a unit-norm viewing direction $\viewdir \in \stwo$. The encoding $\gamma$ is a critical step that maps these coordinates into a higher-dimensional frequency space. The parameters $\params$ are the weights of the continuous MLP, and the output space $\mathcal{S}$ consists of the corresponding volume density $\volumedensity\in [0, 1]$ and directional-emitted color $\mathbf{c}$. In essence, NeRF follows the same idea as~\cite{lombardi2019neural} but adds a radiance component that is view-dependent. The volume density indicates how much radiance (or luminance) is accumulated by a ray passing through $\point$ and is a measure of the effect this point has on the overall scene.  It provides the likelihood that the predicted color value should be taken into account. It can also be interpreted as the differential probability of a ray terminating at an infinitesimal particle at location $\point$. Thus, it can be used to recover the 3D geometry of the scene.

NeRF only requires 2D supervision during optimization. Inspired by earlier neural rendering techniques~\cite{lombardi2019neural}, NeRF leverages differentiable volume rendering to optimize the parameters of an MLP approximating the radiance field via gradient descent, minimizing the photometric error between rendered and observed images. Technically, views of the observed scene are synthesized by casting rays \(\ray(t) = \mathbf{o} + t\viewdir\) from the camera center $\mathbf{o}$, each ray goes through a pixel of the rendered image. To determine the color of pixels, the color accumulated along each ray \(C(\ray)\) is computed via the volume rendering integral:
\begin{equation}
C(\ray) = \int_{t_n}^{t_f} \transmittance(t) \volumedensity(\ray(t)) \thecolor(\ray(t), \viewdir) \, dt,    
\end{equation}
\noi where \(\transmittance(t) = \exp\left(-\int_{t_n}^{t} \volumedensity(ray(t)) \, dt\right)\) is the transmittance, representing the probability of the ray traveling from \(t_n\) to \(t\) without hitting any particles. This integral is approximated via numerical quadrature, typically using stratified sampling, enabling efficient and differentiable rendering. This volume rendering strategy is the key component of the method, allowing the optimization of an accurate 3D representation from 2D images. However, this optimization incurs high computational costs and can take days for big scenes.

\subsubsection{Input encoding}
\begin{figure}[tb]
\centering
    \includegraphics[width=\columnwidth]{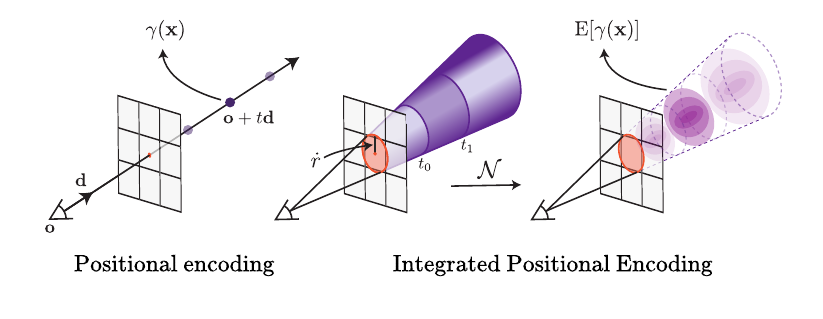}
    \caption{Figure comparing input encoding processes. \textbf{(a)} Original positional encoding~\cite{mildenhall2020nerf}, \textbf{(b)} integrated positional encoding~\cite{barron2021mip}.}
    \label{fig:input_encoding}
\end{figure}

Learning a high-quality mapping from low-dimensional inputs, \eg 3D point coordinates $\point$ and view direction $\viewdir$, to high-dimensional outputs, \eg shape and appearance, critically depends on how the inputs are provided to the neural network (\neuralnetwork). Directly regressing the shape properties $\shapeproperties$ from raw coordinates $\inputparams$ with an MLP~\cite{park2019deepsdf} often fails to reproduce fine-scale variations, as ReLU-based networks inherently prioritize smooth, low-frequency signals. To address this spectral bias, NeRF relies on an input encoding strategy inspired by positional encoding in natural language processing~\cite{vaswani2017attention}, which helps the $\neuralnetwork$ identify the position or scale it is currently processing. 

Following Vaswani~\cite{vaswani2017attention}, Mildenhall \etal~\cite{mildenhall2020nerf} encode the positions $\point\in \rthree$ as a multi-resolution sequence of $m>0$ sine and cosine functions. This \textbf{frequency encoding} represents the $\gamma(\mathbf{q})$ term in Eq.~\ref{eq:3d_representation} (also referred to as \emph{Fourier features mapping}~\cite{tancik2020fourier}). It lifts the low-dimensional queries into a higher-dimensional space, allowing the neural network to better capture high-frequency details in the scene. The hyperparameter $m$ determines the number of frequencies used, controlling both the smoothness and the maximum level of detail captured by the learned representation.

However, standard frequency encoding treats rays as infinitely thin lines, which causes severe aliasing when rendering views at varying resolutions or distances due to the discrete sampling of continuous signals. To address this issue, methods have evolved toward multi-resolution positional encoding. Barron \etal~\cite{barron2021mip} introduced Mip-NeRF, which reasons about the 3D conical frustum defined by a camera pixel rather than a single ray. These frustums are enhanced with Integrated Positional Encoding (IPE), which works by approximating the frustum with a multivariate Gaussian and computing the closed-form integral over the positional encodings within the Gaussian (see Fig.~\ref{fig:input_encoding}). 

Building on this need for robust multi-scale parameterization, recent architectures embed multi-resolution inductive biases directly into the scene representation. While some methods explicitly encode Levels of Detail (LOD) into the latent space~\cite{turki2023pynerf,barron2023zipnerf,VRNeRF} or perform volume integration via convolutions~\cite{hu2023Tri-MipRF,zhuang2023anti} or super-sampling~\cite{barron2023zipnerf}, these approaches often suffer from strict per-LOD supervision requirements or high computational costs. To overcome this, Ring-NeRF~\cite{petit_ring-nerf_2024} introduces a continuous multi-scale grid paired with a ring-based spatial partition.

Point coordinates are equivariant to rotation, translation, and articulated transformations, such as skeletal or non-rigid deformations. Moreover, purely coordinate-based positional encoding requires a fixed number of encoding frequencies and thus does not adapt locally to the scene's complexity. Therefore, instead of directly inferring scene properties from the input $\inputparams$ or its frequency encoding, the neural field can be inferred from learned spatial features or latent codes, along with the network parameters, in an auto-decoding fashion~\cite{park2019deepsdf}. During training, one needs to jointly optimize the network parameters $\params$ and the codes. The main advantage of this representation is that the neural function can now be represented using a small $\MLP$. A popular approach is to precompute local codes, of a single scale~\cite{park2019deepsdf, peng2020convolutional} or multiscale~\cite{muller2022instant}, 
and arrange them in auxiliary data structures such as 3D grids~\cite{peng2020convolutional,sun2022direct,fridovich2022plenoxels,muller2022instant} or trees~\cite{yu2021plenoctrees}. 
The encoding of any point $\domainpoint \in \domain$ is then obtained via trilinear interpolation of the latent codes from the neighboring cell centers.

\subsubsection{Surface-aware density formulation}
\label{subsec:neus}

NeRF is a powerful scene representation that enables novel view synthesis. However, extracting high-quality surfaces from the learned representation is difficult since the density field does not carry sufficient information about the surface geometry.

Methods such as~\cite{yariv2020multiview, wang2021neus, oechsle2021unisurf, long2023neuraludf, liu2023neudf} address this by modeling geometry as a distance function or occupancy field. A first MLP predicts the SDF~\cite{wang2021neus} or occupancy~\cite{oechsle2021unisurf}. Its output is then converted to density using a probability density function $\probabilitydensity_s(f(x))$~\cite{wang2021neus}, where $f(x)$ is the SDF or the occupancy and $\probabilitydensity_s(x)$ is a density distribution, such as the logistic density distribution $\probabilitydensity_s(\point)$:
\begin{equation}
    \probabilitydensity_s(x) = \frac{s e^{-sx}}{1 + e^{-sx}}.
\end{equation}

\noi By introducing this density distribution into neural radiance field representations, the network can learn an implicit surface representation using only volume rendering. This unification results in a more accurate surface representation while preserving robust training—even in the presence of abrupt depth changes—thanks to the smoothing effect of volume rendering. Moreover, the learning process can be further refined by encouraging geometric consistency. This is typically done by incorporating an eikonal term~\cite{wang2021neus}, which enforces that the gradient norm of the learned SDF is equal to one. As a result, the model converges to a well-behaved representation with meaningful surface normals.

\subsubsection{Extracting mesh from implicit representations}
\label{sec:mesh_extraction}

A key challenge to the broad integration of implicit representations in downstream applications is that they are not the standard format for most existing tools and pipelines. As explained in Sec~\ref{sec:discrete_explicit_representations}, polygonal meshes are generally used to infer the geometry precision of reconstructed shapes and are the standard for downstream tasks. Therefore, methods based on implicit representations often need to extract a mesh from the learned representation. 

This is typically done by applying a meshing algorithm (e.g., Marching Cubes~\cite{marching_cubes}, Dual Contouring~\cite{dual_contouring}, Marching Tetrahedra~\cite{doi1991efficient}) as a post‑processing step to obtain a triangle mesh. However, these grid-based sampling methods introduce discretization errors and cannot preserve sharp features unless the grid resolution is excessively high. Furthermore, the resulting mesh is obtained only during post-processing and is not learned jointly with the model. 

Several works~\cite{liao2018deep, gao2020learning, chen2021neural, guillard2022meshudf} integrate this mesh‑extraction process directly into the learning pipeline, effectively turning Marching Cubes or its variants into differentiable network layers. These approaches embed differentiable contouring layers, allowing the network to output vertices and connectivity directly from an implicit representation. While this preserves compatibility with convolutional operations, it often incurs a high memory footprint due to the underlying grid resolution. 

To overcome the precision limits of grid-based sampling, Stippel \etal~\cite{stippel2025marching} recently introduced \textit{Marching Neurons}, an analytical method that extracts the exact zero-level set of a neural implicit function. By exploiting the piecewise-linear structure of ReLU-based networks, it traverses the decision boundaries of neurons directly rather than sampling the space. This allows for the extraction of high-fidelity meshes that faithfully capture the exact geometry encoded by the network weights, eliminating grid aliasing and discretization artifacts without requiring infinite resolution sampling.

\subsection{Optimization and Acceleration}
\label{sec:nerf_optimization}

While neural fields using coordinate-based MLPs ("vanilla" NeRFs) are memory efficient, they suffer from huge computational costs for training and seconds per frame for rendering. This bottleneck is created by the need to query a deep MLP millions of times along rays to resolve volumetric integrals. Recent research has addressed these limitations using methods that can be categorized through five strategies: \textbf{(1) Spatial Subdivision}, breaking the scene into smaller, faster networks; \textbf{(2) Surface Localization}, skipping computation in empty space; \textbf{(3) Baking}, pre-computing static elements into grids; \textbf{(4) Efficient Integration}, replacing sampling with closed-form solutions; and \textbf{(5) Optimizing input encoding}, replacing grids used in input encoding by optimized data structures.

\subsubsection{Acceleration by subdivision}
\label{sec:acceleration_by_subdivision}
\begin{figure}[tb]
\centering
    \includegraphics[width=0.8\columnwidth]{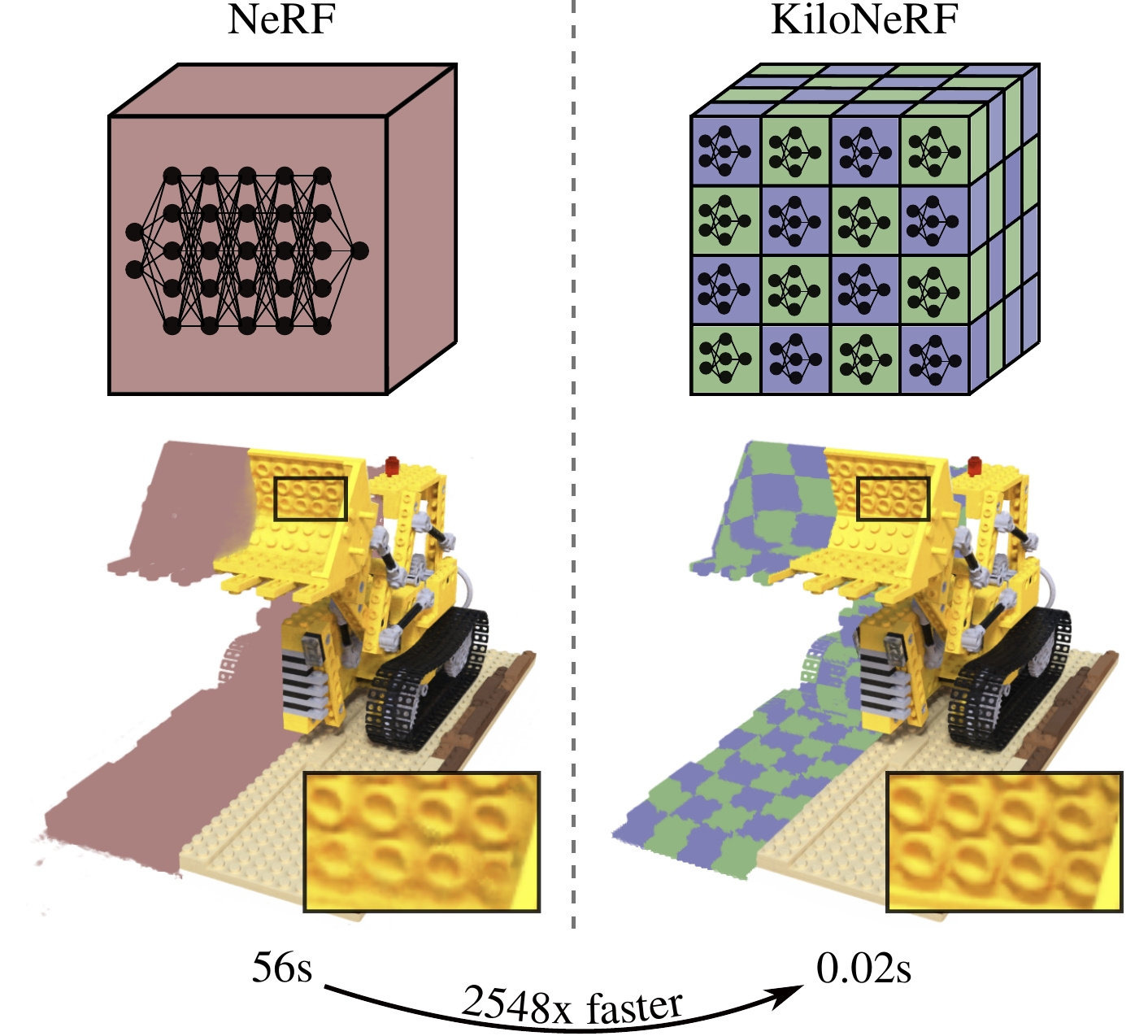}
    \caption{Comparison between vanilla NeRF representation and a subdivided representation that accelerates the rendering. Figure borrowed from~\cite{reiser2021kilonerf}.}. 
    \label{fig:kilonerf}
\end{figure}

One strategy for acceleration is to subdivide a 3D scene into spatial cells and represent the scene properties within each cell using small neural networks that are easier to train and significantly faster to evaluate at runtime; see Fig.~\ref{fig:kilonerf}. The subdivision can take the form of a regular volumetric grid~\cite{reiser2021kilonerf} or irregular Voronoi cells~\cite{rebain2021derf}. The cells can be disjoint or overlapping. The subdivision can also be flat~\cite{reiser2021kilonerf,rebain2021derf} or hierarchical~\cite{liu2020neural,yu2021plenoctrees}.  In some approaches, a single neural network with shared weights is used across all cells~\cite{liu2020neural,yu2021plenoctrees,wang2022fourier}. Others, \eg Reiser \etal~\cite{reiser2021kilonerf}  and Rebain \etal~\cite{rebain2021derf},  train one NeRF per grid cell, each modeled using a tiny MLP instead of the deeper MLP used in the original NeRF formulation. KiloNeRF~\cite{reiser2021kilonerf} adopts a teacher–student training paradigm, where a standard NeRF is first trained, and its knowledge is subsequently distilled into a set of lightweight MLPs arranged in a 3D grid. This strategy results in an acceleration of up to three orders of magnitude compared to the original NeRF.  However, deploying a large number of MLPs incurs a substantial memory overhead.  

\subsubsection{Acceleration by surface localization}
\label{sec:localization}
Standard volumetric rendering is inherently inefficient because it samples the entire ray, even though surface geometry occupies only a fraction of the 3D space. As observed by Piala \etal~\cite{piala2021terminerf} and Hu \etal~\cite{hu2022efficientnerf}, only around $10$ to $20\%$ of the evaluated samples have non-zero density and actually contribute to the final image. To address this, acceleration methods aim to localize the surface interface and concentrate samples only in these relevant regions, effectively skipping empty space. Existing approaches generally fall into four categories based on how they define this region of interest: \textbf{(1)} using sparse data structures~\cite{sun2022direct}, \textbf{(2)} leveraging depth oracles~\cite{neff2021donerf,lin2022efficient}, \textbf{(3)} guiding sampling via Signed Distance Fields~\cite{shao2022doublefield,wan2023learning} or density estimates~\cite{hu2022efficientnerf}, and \textbf{(4)} employing learned sampling networks~\cite{piala2021terminerf,kurz2022adanerf}. By refining the sampling strategy to focus solely on the localized surface, these methods significantly reduce the number of required MLP evaluations, thereby improving computational efficiency without compromising geometric accuracy.

\vspace{6pt}
\noi\textbf{(1) Sparse spatial data structures.}
\label{sec:spatial_data_structures}
When the surface is known a priori, space partitioning techniques that use data structures such as sparse grids~\cite{chabra2020deep,chibane2020implicit,jiang2020local,liu2020neural,peng2020convolutional},  octrees~\cite{li2020monocular,takikawa2021neural,yu2021plenoctrees,wang2022fourier,hu2022efficientnerf}, and VDBs~\cite{hyan2023plenvdb},  can be used to localize the surface and eliminate empty voxel cells. This ensures that neural fields are only evaluated in non-empty regions. The main challenge, however, is that in a neural field setting, surfaces typically emerge during training. Thus, one cannot precompute space-partitioning data structures in advance.

To address this issue, some methods learn implicit functions over an initial voxel grid, subdividing the 3D space~\cite{liu2020neural,li2020monocular, sun2022direct}. This produces a coarse approximation of the scene geometry. They then progressively prune empty voxels during training based on this initial geometry signal. This pruning strategy, whether it uses a sparse grid~\cite{liu2020neural} or a hierarchical octree~\cite{li2020monocular},  accelerates the surface localization while enabling progressive training by dynamically adapting the voxelization to the underlying scene structure. The use of a sparse voxel grid allows the renderer to skip empty regions during tracing a ray,  leading to rendering speedups of up to $10$ times compared to the standard NeRF implementation.

TNSVF~\cite{liu2020neural} and several other  works~\cite{sun2022direct,yu2021plenoctrees}  adopt a multistage, coarse-to-fine strategy in which regions are progressively refined and pruned as necessary.  EfficientNeRF~\cite{hu2022efficientnerf}, on the other hand,  builds a two-level octree composed of a coarse density voxel volume used to cache the approximate geometry (in the form of densities), and fine sparse voxel grids localized around the surface. The former is used to localize the object's surface, while the latter is used to refine geometry and radiance estimation in surface-adjacent regions. This results in a data structure called NerfTree, which can be seen as an Octree but only with two levels.  The density voxel volumes are used during training and testing. They are progressively updated during the training process. As a result, the approach reduces the training time by up to $88\%$ compared to the traditional NeRF, and improves the inference speed, rendering images at more than $200$ fps. However, this comes at the cost of a larger memory footprint. Despite the acceleration, this method introduces an unavoidable trade-off between memory usage and precision arising from the voxel structure and the density grid. 

Hyan \etal~\cite{hyan2023plenvdb} take a different approach by directly learning a VDB, a hierarchical data structure for sparse volumes~\cite{kim2022neuralvdb},  from a set of posed images. They then use this representation for real-time rendering. These methods are inspired by classical computer graphics techniques such as the Bounding Volume Hierarchy (BVH)~\cite{rubin19803} and the Sparse Voxel Octree (SVO)~\cite{laine2010efficient}, which model a scene in a sparse hierarchical form to accelerate ray tracing. This technique achieves $10\times$ to $20\times$ speed-up over traditional NeRF, with a rendering quality comparable to Neural Volumes~\cite{lombardi2019neural} and Scene Representation Networks~\cite{sitzmann2019scene}.

Note that while these techniques are effective, they lead to a more complex training process, in which the sparse data structure must be periodically updated during training~\cite{muller2022instant} because surfaces emerge only progressively.   

\vspace{6pt}
\noi\textbf{(2) Using depth oracles.} DoNeRF~\cite{neff2021donerf} swaps the coarse network of the original NeRF with a depth oracle network, which provides suitable sample locations for the second shading network, reducing the sample counts per ray by up to $128\times$. However, this architecture is not end-to-end trainable and struggles without high-quality depth supervision.  Inspired by deep learning techniques for multiview stereo~\cite{laga2020survey}, ENeRF~\cite{lin2022efficient} first builds a cascade cost volume, also referred to as a 3D feature volume, and uses it to predict a coarse geometry of the scene in the form of depth and confidence maps. The coarse geometry is then used to sample a few points near the scene surface, significantly improving the rendering speed,  while still achieving competitive accuracy. The 3D feature volume also provides rich geometry-aware features for constructing a generalizable radiance field.

\vspace{6pt}
\noi\textbf{(3) Using signed distance fields.} Hu \etal~\cite{hu2022efficientnerf} observed that a large number of 3D points have zero density and thus do not contribute to the geometry and radiance estimation. They proposed evaluating the coarse NeRF MLP only at valid points, \ie, points with positive densities, while the fine network is evaluated at pivotal points, defined as those whose densities exceed a predefined threshold.  Shao \etal~\cite{shao2022doublefield} use the SDF field to guide surface sampling. They estimate ray-surface intersections using the SDF and then perform fine-grained sampling around the intersection point. Similarly, Wan \etal~\cite{wan2023learning} use the volume density or SDF estimates from the implicit representation to localize the surface between an inner and an outer mesh. Learned features are attached to the mesh vertices. Given a viewing direction $\viewdir$, the method computes the ray intersection with both meshes, interpolates features at the intersection points, and feeds them, along with the viewing direction, into a shallow convolutional network to estimate radiance. Unlike previous methods, this approach requires only two samples per ray, significantly improving runtime efficiency.  Moreover, it is general and can be applied as a post-processing step to any existing NeRF models.

\vspace{6pt}
\noi\textbf{(4) Sampling networks.} Some methods, \eg TermiNeRF~\cite{piala2021terminerf} and AdaNeRF~\cite{kurz2022adanerf},  use  a sampling network  to localize surfaces. These methods split the NeRF network into two parts, a sampling network and a shading network, which are trained jointly. The former predicts suitable sample locations with a single evaluation per view ray, while the latter adaptively shades only the most significant samples per ray. TermiNeRF~\cite{piala2021terminerf} conditions the sampling network on the density of a pretrained NeRF. It uses the whole range of samples without the need for a depth map. This improves quality in geometrically ambiguous regions. While TermiNeRF can be trained end-to-end, it requires a  pre-trained NeRF to initialize the color network to achieve the best results. AdaNerf \etal~\cite{kurz2022adanerf} employs fixed sample positions along each ray and incrementally introduces sparsity during training. After fine-tuning to the desired number of samples, the resulting compact neural representation can be rendered in real-time. The approach is adaptable and can be fine-tuned to meet performance targets for real-time applications.

\begin{figure}[tb]
    \centering
    \includegraphics[width=\columnwidth]{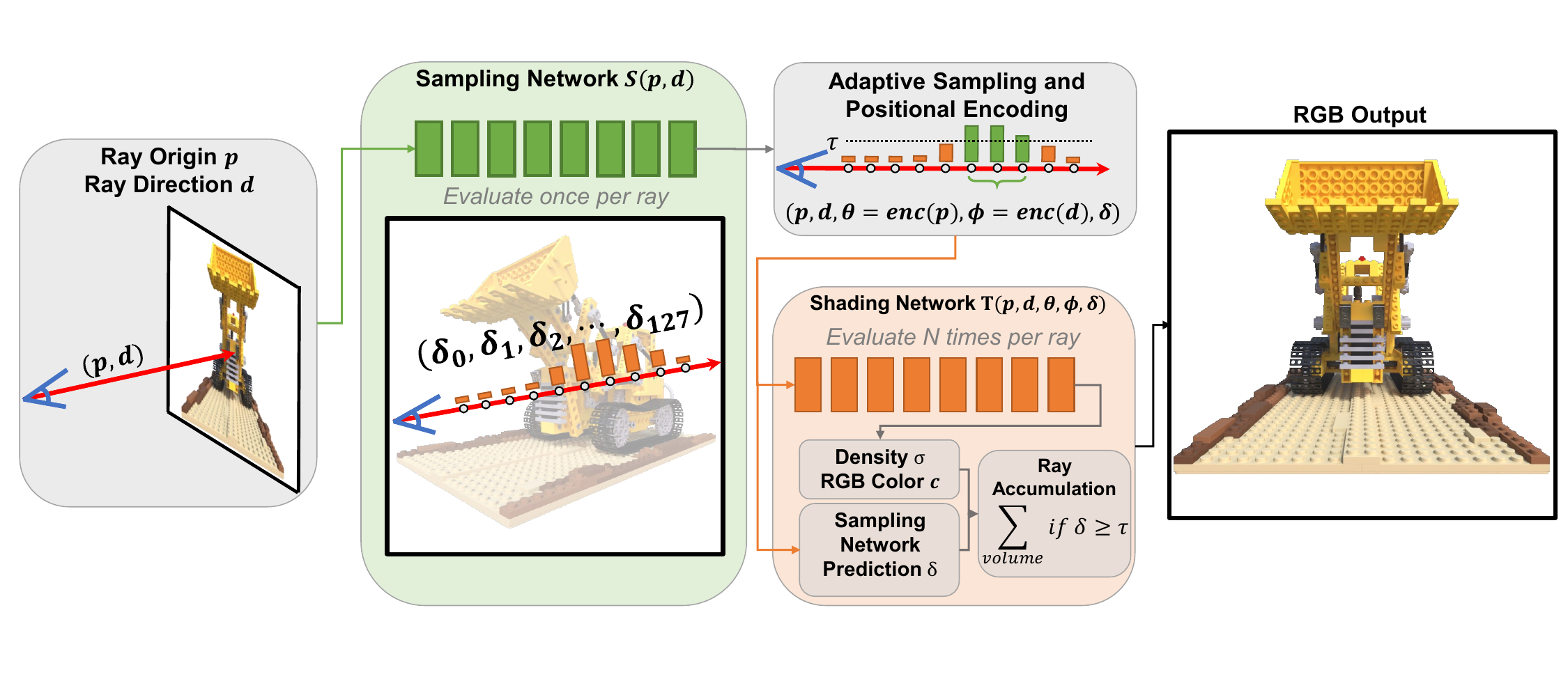}
    \caption{Overview of a typical pipeline using a sampling network. To accelerate rendering, the architecture splits the traditional MLP into a lightweight Sampling Network that evaluates once per ray to predict sample importance weights, and a heavier Shading Network that only processes the most significant. Image borrowed from~\cite{kurz2022adanerf}.}
    \label{fig:adanerf_pipeline}
\end{figure}

\vspace{6pt}
\noi\textbf{(4) Sampling networks.} Some methods~\cite{piala2021terminerf,kurz2022adanerf}, use a sampling network(see Fig.~\ref{fig:adanerf_pipeline})to localize surfaces. These methods split the NeRF network into two parts, a sampling network and a shading network, which are trained jointly. The former predicts suitable sample locations with a single evaluation per view ray, while the latter adaptively shades only the most significant samples per ray. TermiNeRF~\cite{piala2021terminerf} conditions the sampling network on the density of a pretrained NeRF. It uses the whole range of samples without the need for a depth map. This improves quality in geometrically ambiguous regions. While TermiNeRF can be trained end-to-end, it requires a pre-trained NeRF to initialize the color network to achieve the best results. Kurz \etal~\cite{kurz2022adanerf} employs fixed sample positions along each ray and incrementally introduces sparsity during training. After fine-tuning to the desired number of samples, the resulting compact neural representation can be rendered in real-time. The approach is adaptable and can be fine-tuned to meet performance targets for real-time applications.

\subsubsection{Acceleration by baking neural fields}
\label{sec:precomputing}

Another approach to speeding up neural fields at runtime is by precomputing data that remains static during inference. This process is referred to as \textit{baking neural fields}. For instance, radiance, and thus color, is typically composed of a view-independent diffuse component and a view-dependent specular component. The former remains constant regardless of view, while the latter varies with the camera angle. Thus, several papers~\cite{hedman2021baking,yu2021plenoctrees,wang2022fourier} have exploited this decomposition to accelerate NeRF rendering by precomputing and storing the view-independent part in a sparse 3D structure and deferring view-dependent computations to runtime. 

For instance, Hedman \etal~\cite{hedman2021baking} reformulate NeRF so that it outputs, in addition to the volume density, the diffuse RGB color and view-dependent 4D feature vectors. To render a pixel, the method accumulates the diffuse colors and the feature vectors along a ray, concatenates them with the view direction, and passes them to a shallow MLP to produce a view-dependent residual, which is then added to the diffuse color.  This structure enables precomputing and storing the diffuse colors and the 4D feature vectors within a sparse voxel grid representation.  The resulting scene representation retains the ability to render fine geometric details and view-dependent appearance, is compact (averaging less than $90$ MB per scene), and can be rendered in real-time (higher than $30$ fps on a laptop GPU for images of size $800\times 800$). Also, the baked information is stored in a texture atlas. Thus, image compression algorithms can be used to reduce the storage size. However, this compression leads to relatively low-quality metrics. 

FastNeRF~\cite{garbin2021fastnerf} further reduces runtime cost by caching the view-independent component of the radiance. This is done by precomputing radiance values at selected 3D points, which are sorted in a sparse spatial structure, \eg voxel grid vertices.  These cached values can be retrieved much faster than evaluating them on the fly with a full NeRF. FastNeRF can render high-fidelity photorealistic images at $200$fps on a high-end consumer GPU. However, caching the radiance values into explicit spatial data structures comes at the expense of model compactness.

Yu \etal~\cite{yu2021plenoctrees} introduced NeRF-SH, which achieves this factorization of appearance via closed-form spherical harmonic basis functions. The method approximates the view-dependent color —defined as a function of a point $\point\in \rthree$ and viewing direction $\viewdir$— as a weighted sum of spherical harmonic bases. The key idea is to define the weights as a function of the point $\point$ and the harmonic basis functions as functions of the viewing direction. This factorization allows NeRF to be reformulated such that the MLP regresses only the view-independent information—\ie the density and the coefficients of the spherical harmonic basis—which are then stored in a PlenOctree structure to avoid sampling in empty regions.  Thus, at training, the network does not require the viewing direction as input. During inference, the view-dependent color at a given point $\point$ along a ray $\viewdir$ is determined by only querying the spherical harmonic functions. This results in a substantial inference-time speed-up:  the approach accelerates the rendering performance of the original NeRF method by more than $3000$ times, achieving a frame rate above $150$ fps for images of resolution $800\times800$, while producing images that are equal to or better in quality than NeRF. However, training remains time-intensive, typically requiring one or two days for convergence.

PlenOctrees~\cite{yu2021plenoctrees} has since been extended in various ways to mitigate the limitations of the original method. For example, the acquisition of spherical harmonic coefficients and densities remains time-consuming. Wang \etal~\cite{wang2022fourier} addressed this issue by using a coarse-to-fine fusion scheme that leverages the generalizable NeRF technique~\cite{chen2021mvsnerf,wang2021ibutter,wang2021ibrnet}   to generate the partitioning tree via spatial blending. Additionally, to handle dynamic scenes, the approach tailors the implicit network to model the Fourier coefficients of time-varying density and color attributes. 

PlenOctrees~\cite{yu2021plenoctrees}  assumes that the opacity and spherical harmonic coefficients remain constant inside each voxel. Plenoxel~\cite{fridovich2022plenoxels} relaxes this assumption by using trilinear interpolation to define a continuous function throughout the volume.  Furthermore, unlike PlenOctrees~\cite{yu2021plenoctrees}, Plenoxel~\cite{fridovich2022plenoxels} directly optimizes voxel opacities and spherical harmonic coefficients without using neural networks.

\subsubsection{Accelerating the integration} 
\label{sec:nerf_integration}

For each pixel, NeRF requires sampling hundreds of points along the corresponding ray during both training and inference. At each sample point, the network is queried for density and color, and these values are accumulated via volume rendering to compute the final pixel color. Monte Carlo integration is typically used to approximate this process, but it involves millions of rays, each requiring hundreds of forward passes through a neural network, resulting in a significant computational bottleneck that hinders the practical deployment of NeRF.

To address this issue, Lindell \etal~\cite{lindell2021autoint} proposed AutoInt,  an auto-integration mechanism that learns efficient, closed-form solutions to integrals using coordinate-based neural networks. The underlying idea is based on the observation that taking the derivative of a coordinate-based network results in a new computational graph, a \emph{grad network}, which shares the parameters of the original network. In other words, taking the derivative of an MLP results in a gradient network that can be trained on a signal that one wishes to integrate. By reassembling the gradient network parameters back into the original MLP, one can construct a neural network that represents the anti-derivative of the signal. This procedure, referred to as AutotInt, results in a closed-form solution for the anti-derivative, which, by the fundamental theorem of calculus, enables the calculation of any definite integral in two evaluations of the MLP. Thus, the approach allows for the computation of integrals efficiently and automatically, without relying on traditional numerical techniques, such as sampling or finite differences. As a result, this accelerates the rendering in neural fields.

In contrast, Wu \etal~\cite{wu2022diver} and later Rivas \etal~\cite{rivas2023nerflight} proposed deterministic integration techniques that approximate the volume rendering integral using a decoder MLP with learnable parameters. In these methods, each ray is decomposed into segments corresponding to the voxels it intersects. A decoder network estimates the contribution of each voxel segment, and the accumulated results form the final pixel color. These methods offer improved efficiency and are particularly effective at capturing thin translucent structures, which are often missed by conventional integrators.

\subsubsection{Optimizing the feature grids}
\label{sec:feature_grids}

\begin{figure}[tb]
\centering
    \includegraphics[width=\columnwidth]{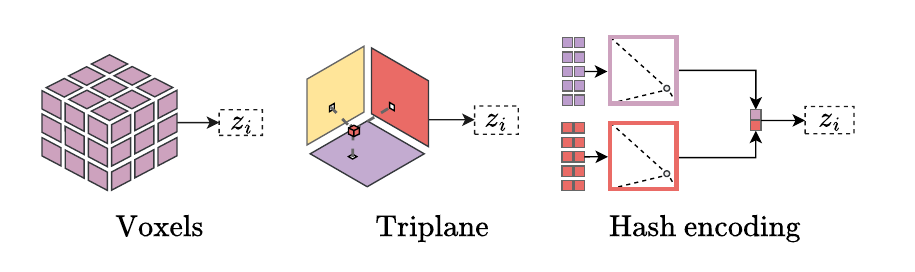}
    \caption{Comparison of Spatial Feature Encoding Strategies. (a) Voxel Grids store learnable features. (b) Tri-planes decompose the 3D volume into three orthogonal 2D feature maps. (c) Multiresolution Hash Encoding maps spatial coordinates at multiple resolution levels to a compact hash table of learnable feature vectors.}
    \label{fig:feature_encoding}
\end{figure}

Accurately representing the geometry and appearance of a 3D scene requires high-resolution feature grids. However, most of the features in such grids correspond to empty regions of the 3D space. As a result, several methods investigated efficient arrangements of the learned features, which are visualized in Fig~\ref{fig:feature_encoding}. This is achieved using:

\vspace{6pt}
\noi\textbf{(1) Sparse data structures.} Sparse data structures such as Octrees~\cite{takikawa2021neural} allow for coarsely sampling empty regions and densely sampling regions near the surface. While efficient, Octree-based methods still require large storage and GPU memory. They also require a priori knowledge of the geometry of the scene. 

\vspace{6pt}
\noi\textbf{(2) 2D grids of features.} Instead of computing volumetric feature grids, some methods arrange the features into planar grids. Popular methods include triplane~\cite{chan2022efficient}, which uses three orthogonal planes aligned with the coordinate system. The feature at a given point $\point$ is obtained by first projecting the point onto each of the three planes, interpolating the features of the neighboring cells, and finally aggregating the three features either by concatenation or pooling. Other methods use a single 2D ground plane, which can be defined at a single or multiple resolutions~\cite{xu2023grid}. These methods are particularly well-suited for representing large-scale environments, such as city-scale scenes, as well as aerial imagery~\cite{xu2023grid}.

\vspace{6pt}
\noi\textbf{(3) Arranging features in texture maps.} To accelerate the rendering, Chen \etal~\cite{chen2023mobilenerf}  and later Gu \etal~\cite{gu2024ue4} represent a scene as a polygonal mesh, whose texture map stores features and opacity values. During rendering,  given a camera pose, Chen \etal~\cite{chen2023mobilenerf} adopt a two-stage deferred rendering process. \First,  the mesh is rasterized to the screen space and a feature image is constructed, creating a deferred rendering buffer in GPU memory. \Second, these features are converted into a color image via a (neural) deferred renderer running in a fragment shader, \ie a small MLP, which receives a feature vector and a view direction and outputs a pixel color. This method achieves interactive frame rates on a wide range of compute platforms, including mobile phones. The approach fully exploits the parallelism of modern graphics hardware, including z-buffering and fragment shaders, achieving $10\times$ faster rendering than SNeRG~\cite{hedman2021baking} with comparable output quality on standard benchmark scenes.

\noi \textbf{(4) Caching and hashing}. Instead of reducing the number of features, other methods use caching and hashing. For example, Muller \etal~\cite{muller2022instant} store the trainable feature vectors used in feature grid-based representations, in compact multiresolution spatial hash tables  (one per resolution), which are then encoded using small MLPs. They also apply half-precision for feature storage, improving the compactness of the model. This representation does not require progressive pruning during training, nor does it rely on prior knowledge of the scene geometry. The multiresolution structure helps disambiguate hash collisions, resulting in a parallelizable architecture well-suited for modern GPUs. This allows learning and rendering a scene representation in real-time. The approach, however, is still unclear on how to optimally tune its hash table size to remain as compact as fully neural representations. Also, the multiple linear interpolations needed due to the multiresolution hash grid lead to a reduction in rendering speed. 

\noi \textbf{(5) Shared feature grids}. Instead of using a large (multi-resolution) feature grid, NerfLight~\cite{rivas2023nerflight}  partitions the 3D volume into regions and uses small MLPs, one per region, to decode the radiance field. The key feature is that the MLPs share the same feature grid. This results in a smaller grid where each feature is located in more than one spatial position, forcing them to learn a compact representation that is valid for different parts of the scene. NerfLight~\cite{rivas2023nerflight} disposes the features symmetrically on each region, which favors feature pruning after training while also allowing smooth gradient transitions between neighboring voxels. This leads to high-quality, seamless reconstruction and even faster and lighter models.

\noi \textbf{(6) Tensor decomposition and vector quantization}. Instead of storing the volumetric feature grids, TensoRF~\cite{chen2022tensorf} factorizes them into compact components, leading to significantly higher memory efficiency. Formally, TensoRF~\cite{chen2022tensorf}  learns two feature grids: one for the geometry and another for appearance. The former can be treated as a 3D tensor, while the latter can be treated as a 4D tensor. The idea is then to use decomposition methods to factorize the feature volumes into compact components. 

A popular decomposition method is the classic CANDECOM/PARAFAC (CP) decomposition~\cite{carroll1970analysis}. Given a 3D tensor $\tensor \in \real^{I\times J \times K} $, the classic CP decomposition factorizes it into a sum of outer products of vectors:
\begin{equation}
    \tensor = \sum_{r=1}^R \vertex_r^1 \outerproduct \vertex_r^2 \outerproduct \vertex_r^3,
\end{equation}

\noi where $\vertex_r^1 \outerproduct \vertex_r^2 \outerproduct \vertex_r^3$ corresponds to a rank-one tensor component, and $\vertex_r^1 \in  \real^{I}$, $\vertex_r^2  \in  \real^{J}$, and  $\vertex_r^3 \in \real^{K}$ are factorized vectors of the three modes of the $r$ component. While this method can be directly applied to radiance field modeling, due to its high compactness, it can result in many components to model complex scenes. This can lead to high computational costs. 

Instead of using CP decomposition, TensoRF~\cite{chen2022tensorf} proposed Vector-Matrix decomposition, which factorizes a tensor (feature grids in our case) into multiple vector-matrix outer products. This factorization allows the computation of each voxel's feature vector at low cost, only requiring one value per XYZ-mode vector/matrix factor. This significantly reduces model size: it brings the memory requirement of feature grid-based NeRF to the same level as point encoding-based NeRF while significantly reducing the training time. 

VQAD~\cite{takikawa2022variable}  presents a dictionary method for compressing such feature grids, reducing their memory consumption by up to $100x$ and permitting a multiresolution representation, which can be useful for out-of-core streaming. VQAD~\cite{takikawa2022variable} formulates the dictionary optimization as a vector-quantized auto-decoder problem, which enables learning end-to-end discrete neural representations in a space where no direct supervision is available and with dynamic topology and structure.

\noi \textbf{(7) Feature grids $+$ ReLU}.
Several other mechanisms have been proposed to accelerate neural fields. For instance, Kharnewar \etal~\cite{karnewar2022relu} showed that by simply using ReLU non-linearity on top of interpolated grid values, without any additional learned parameters,  combined with coarse-to-fine optimization, feature grid-based methods become competitive with the state-of-the-art coordinate-based MLPs.

\subsection{Differentiable Primitive Splatting}
Recent advances in view synthesis have shifted towards volumetric representations that combine the speed of primitive-based rasterization~\cite{Zwicker_2001_EWA, lassner2021pulsar} with the differentiable optimization of Neural Radiance Fields~\cite{mildenhall2020nerf}.

This paradigm, which we refer to as \textit{Differentiable Primitive Splatting}, approximates the scene as a collection of disjoint or overlapping geometric primitives. Kerbl \etal~\cite{Kerbl_2023_3dGS} pioneered this approach by representing the scene using a set of anisotropic 3D Gaussians, which are then projected (or splatted) onto the image plane to synthesize novel views. The proposed pipeline has then been extended to other primitive shapes and density functions~\cite{hamdi_ges_2024,Mai_2025_ICCV,Huang2DGS2024,held_3d_2025,liu_deformable_2025,held_3d_2025}.

This section details the 3D Gaussian Splatting(3DGS) method as the foundational implementation of this paradigm. We first detail the parameterization of the primitives and the splatting process (Section~\ref{sec:3dgs}). We then discuss improvements of the original method (Section~\ref{sec:3dgs_optimization}). Finally, we discuss generalizations of this method to other primitive shapes and density functions (Section~\ref{sec:alt_primitives}).

\subsubsection{3D Gaussian Splatting}\label{sec:3dgs}

\begin{figure}[t!]
\centering
    \includegraphics[width=\columnwidth]{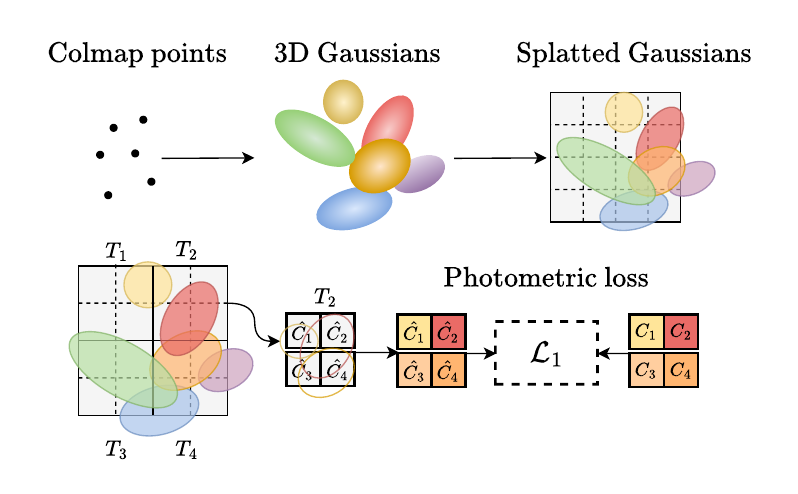}
    \caption{Depicting the 3DGS~\cite{Kerbl_2023_3dGS} pipeline. 3D Gaussians are initialized from a sparse point cloud obtained from SfM (Colmap~\cite{schoenberger2016sfm}), and they are then projected (splatted) into the 2D image space. The image plane is then divided into non-overlapping patches (tiles). Gaussians are replicated as needed, if they span multiple tiles, and then sorted by depth. Pixel colors are computed using $\alpha$-blending. The final rendering image is compared to the ground truth image using a photometric loss function.}
    \label{fig:3dgs}
\end{figure}

Similarly to NeRF~\cite{mildenhall2020nerf}, 3D Gaussian Splatting (3DGS)~\cite{Kerbl_2023_3dGS} learns a volumetric radiance field from a set of posed images by optimizing a differentiable rendering pipeline. However, instead of using a continuous MLP to represent the radiance field, 3DGS approximates it using a set $\gaussians = \{g_1, g_2, \dots, g_N\}$ of anisotropic 3D Gaussians. Under our formulation (Eq.~\ref{eq:3d_representation}), the query $\mathbf{q} = \point$ is a continuous 3D spatial coordinate. The representation parameters $\params$ are explicitly stored as the attributes of the Gaussians (means, covariance matrices, opacities, and spherical harmonics). The mapping function $\neuralsurfacefunc_\params$ aggregates the influence of these primitives to evaluate the final volumetric properties $\mathcal{S}$ at the queried location. Each primitive $g_i$ is defined by a covariance matrix $\cov_i$ centered at mean $\mu_i$:

\begin{equation}
  g_i(\point) = e^{-\frac{1}{2}\point^T \cov_i^{-1} \point}
\end{equation}

\noi where $\point \in \rthree$ is a 3D point in space. Each Gaussian $g_i$ also encodes appearance attributes, including opacity $\alpha_i$ and view-dependent color $\thecolor_i$.
While a Gaussian distribution theoretically has infinite extent, its probabilistic density is centered at $\mu_i$ and decays according to $\cov_i$. However, to efficiently represent 3D, each primitive must maintain a physical extent. With Gaussians, this is true only if the covariance matrix is positive definite. Optimizing the parameters of the covariance matrix directly through gradient descent does not guarantee this property. Therefore, in 3DGS, the covariance is implemented as the configuration of an ellipsoid rotated and scaled into world space:
\begin{equation}
    \cov_i = \mathbf{R}_i\mathbf{S}_i\mathbf{S}_i^\top\mathbf{R}_i^\top
\end{equation}
where $\mathbf{s}_i \in \mathbb{R}^3_+$ is a scaling vector and $\mathbf{R}_i$ is a rotation matrix obtained from a normalized quaternion $\mathbf{q}_i$. This decomposition ensures that $\cov_i$ is valid by construction while allowing independent control over the primitive's orientation and size.
To learn the scene representation, 3DGS optimizes the parameters of $\gaussians$ by rendering novel views and minimizing a photometric loss against ground truth images. Unlike ray-marching approaches, 3DGS utilizes a tile-based rasterization pipeline, depicted in Fig~\ref{fig:3dgs}, which consists of three stages:

\begin{enumerate}
    \item \textbf{3D-to-2D Projection:} 
    Based on the EWA splatting formulation~\cite{Zwicker_2001_EWA}, the 3D Gaussians are projected into the 2D image plane. The local affine approximation of the viewing transformation allows the 2D covariance $\cov'_i$ to be computed as:
    \begin{equation}
        \cov'_i = \mathbf{J}\mathbf{W}\cov_i\mathbf{W}^\top\mathbf{J}^\top,
    \end{equation}
    where $\mathbf{W}$ is the viewing transformation matrix and $\mathbf{J}$ is the Jacobian of the affine approximation of the projective transformation.
    
    \item \textbf{Tile-Based Sorting:} 
    In this stage, images are partitioned into $16\times16$ pixel tiles. Gaussians intersecting the view frustum are assigned to the tiles they overlap. To enable correct alpha compositing, these instances are sorted by depth (and tile ID) using a GPU-optimized Radix sort~\cite{Merrill2010}.

    \item \textbf{Alpha Compositing:} 
    For every pixel, the color $C$ is computed by traversing the sorted list of overlapping Gaussians and accumulating their contribution via standard $\alpha$-blending:
    \begin{equation}
        C = \sum_{i \in \mathcal{N}} \mathbf{c}_i \sigma_i \prod_{j=1}^{i-1}(1 - \sigma_j),
    \end{equation}
    where the effective blending weight $\sigma_i$ is given by multiplying the learned opacity $\alpha_i$ with the 2D Gaussian evaluation $g'_i(\mathbf{x})$ at pixel $\mathbf{x}$.
\end{enumerate}

The number of Gaussians $N$ is not fixed. They are initialized from sparse SfM points, which are insufficient to represent complex geometry. Hence, 3DGS dynamically adjusts the number of primitives during optimization. This \textit{Adaptive Density Control} identifies regions of under-reconstruction (missing geometry) or over-reconstruction (coarse approximations) by monitoring the view-space positional gradients of the primitives. Under-represented areas are densified: small primitives are cloned to fill space, while large primitives are split into smaller ones. Simultaneously, a pruning mechanism periodically removes Gaussians from over-represented areas, ensuring the representation remains compact.

This pipeline enables real-time rendering through massive GPU parallelism while maintaining full differentiability, which allows the optimization of Gaussian parameters. The Gaussian representation avoids costly neural network queries required by neural approaches, enabling real-time rendering at high resolutions and fast convergence during training.

\subsubsection{Optimizing 3D Gaussian Splatting}\label{sec:3dgs_optimization}

\begin{table}[t!]
\centering
\scriptsize
\renewcommand{\arraystretch}{1.3}
\setlength{\tabcolsep}{3pt}

\newcommand{\perf}[1]{\textcolor{teal}{#1}}     
\newcommand{\avg}[1]{\textcolor{orange}{#1}}    
\newcommand{\bad}[1]{\textcolor{red!70!black}{#1}} 

\begin{tabularx}{\columnwidth}{l c c c X l}
\toprule
\textbf{Goal} & 
\textbf{Mem.} & 
\textbf{Geom.} & 
\textbf{Artif.} & 
\textbf{Core Mechanism} & 
\textbf{Key Ref.} \\
\midrule

\textbf{Pruning} & 
\high & 
\midlow & 
\midlow & 
Remove redundant/invisible Gaussians &
\tiny{\cite{papantonakis2024reducing, lee2024compact}} \\

\textbf{Compression} & 
\high & 
\midlow & 
\midlow & 
Vector Quantization / Hash-grids &
\tiny{\cite{niedermayr2024compressed, Morgenstern_2023_Compact}} \\

\textbf{Anti-Aliasing} & 
\midlow & 
\midlow & 
\high & 
3D smoothing / Mip-filters &
\tiny{\cite{Yu2023MipSplatting, Yan_2024_CVPR}} \\

\textbf{Artifact Removal} & 
\midlow & 
\midlow & 
\high & 
Per-pixel sort / Ray-tracing (Fixes Popping) &
\tiny{\cite{radl2024stopthepop, yu2024gaussian}} \\

\textbf{Geometry Align.} & 
\midlow & 
\high & 
\midlow & 
SDF / Depth / Normal priors &
\tiny{\cite{guedon2024sugar, yu2024gsdf, li2024dngaussian}} \\

\bottomrule
\end{tabularx}
\caption{\textbf{3DGS Optimization Landscape.} 
Comparison of different optimization targets.
\textbf{Mem}: Storage efficiency, \textbf{Geom}: Underlying geometric quality, \textbf{Artif.}: Visual artifact correction (e.g., aliasing, popping).
\emph{Legend:} \perf{$\bullet$}~Significant Improvement, \avg{$\circ$}~Neutral/No major change.}
\label{tab:3dgs_optimization}
\end{table}

Although 3DGS achieves impressive real-time rendering performance, the original method suffers from three primary limitations: excessive memory consumption due to millions of primitives, rendering artifacts (aliasing and temporal instability), and uncertain geometric underlying structure. This section reviews strategies,  compared in Tab~\ref{tab:3dgs_optimization} designed to mitigate these issues.

\vspace{6pt}
\noi\textbf{(1)Memory efficiency.}
Densification of primitives in 3DGS often leads to millions of redundant primitives, creating a large memory footprint. 
To address this, recent works focus on pruning unnecessary Gaussians. Strategies include ranking Gaussians by a spatial redundancy score~\cite{papantonakis2024reducing} or employing volume-based masking~\cite{lee2024compact} to continuously cull unnecessary primitives during optimization.
Parallel efforts target the compression of Gaussian attributes. Since view-dependent color (Spherical Harmonics) accounts for the greater part of the storage, methods propose reducing the number of SH bands adaptively based on color variance~\cite{papantonakis2024reducing} or replacing them entirely with hash-grids and shallow neural fields~\cite{Morgenstern_2023_Compact}.
Finally, parameters like rotation and scale can be compressed via quantization. They can be stored in compact codebooks obtained through K-means clustering~\cite{papantonakis2024reducing, niedermayr2024compressed} or Residual Vector Quantization~\cite{lee2024compact}. Resulting codebooks can further be compacted by quantization-aware fine-tuning and entropy encoding.

\vspace{6pt}
\noi\textbf{(2) Rendering Fidelity and Artifact Suppression.}
The discrete nature of 3DGS representation introduces two major sources of artifacts: aliasing (spatial) and popping (temporal).
\textit{Aliasing} occurs because 3DGS samples the scene at pixel centers, ignoring the footprint of the pixel itself. When changing resolution (zooming), this causes aliasing. Mip-Splatting~\cite{Yu2023MipSplatting} addresses this by applying a 2D rendering filter that mimics the physical box filter of a camera sensor, effectively band-limiting the signal. Orthogonal multi-scale approaches~\cite{Yan_2024_CVPR, zhang2024fregs} introduce frequency regularization to ensure consistent appearance across resolutions.
\textit{Popping artifacts} arise from the depth-sorting step. As the camera moves, two overlapping Gaussians may swap their sort order, causing a sudden discontinuity in the alpha-blending accumulation. To resolve this, Radl \etal~\cite{radl2024stopthepop} replaces global sorting with a per-pixel depth comparison, while other methods adopt hybrid transparency blending~\cite{hahlbohm2025efficient} or integrate ray-tracing kernels~\cite{yu2024gaussian, moenne20243d} to ensure perspectively accurate compositing.

\vspace{6pt}
\noi\textbf{(3) Improving geometry.} Original 3DGS optimizes for photometric loss, often resulting in unstructured density clouds that do not represent a valid surface. To recover high-quality geometry, recent methods impose physical constraints during optimization. This can be achieved by integrating Signed Distance Fields (SDFs) into the 3DGS framework~\cite{yu2024gsdf, guedon2024sugar}.  Specifically, a regularization term can be introduced to minimize the difference between an ideal SDF and the SDF calculated from Gaussians~\cite{guedon2024sugar}. Alternatively, a higher geometrical precision can be obtained by using a dual-branch framework with GS- and SDF-branches, jointly optimized to provide mutual guidance~\cite{yu2024gsdf}. 
Other methods propose to incorporate geometric regularization, such as monocular depth priors~\cite{li2024dngaussian, Chung_2024_CVPR} to improve geometry representation. However, strict depth constraints may lead to incorrect shifts in Gaussian positions. Therefore, methods further improve representations by using multi-plane Gaussian optimization, multi-scale geometric correction~\cite{Li_2025_mpgs}, or multiview geometric constraints~\cite{Kim_2025_multiview}.

\subsubsection{Using different primitives}\label{sec:alt_primitives}
While 3D Gaussians offer mathematical convenience, their symmetric, unbounded nature limits their ability to model hard surfaces, sharp edges, or complex topologies. Consequently, recent works have explored alternative primitives to improve geometric fidelity and rendering efficiency.

To better capture high-frequency details and reduce artifacts, several methods use alternative density functions. Hamdi \etal~\cite{hamdi_ges_2024} proposed using Generalized Exponential Functions coupled with a frequency-modulated image loss to better capture sharp details and reduce the memory footprint. Similarly, Liu \etal~\cite{liu_deformable_2025} introduced Beta Kernels, which offer bounded support and adaptive frequency control, capturing fine geometric details with higher fidelity while improving memory efficiency. Addressing the softness of the density field, Mai \etal~\cite{Mai_2025_ICCV} introduced ellipsoids with constant density and analytically integrated optical depth (EVER) to resolve blurring artifacts.

Other approaches prioritize explicit geometric alignment by leveraging 2D primitives. Huang \etal~\cite{Huang2DGS2024} proposed 2D Gaussian Splatting (2DGS), which replaces 3D Gaussians with oriented 2D disks (surfels) that align with the local surface tangent plane. This surface-centric approach is further enhanced by deformable radial kernels~\cite{huang_deformable_2025}, which allow adaptive shaping of the splatting kernels to capture intricate details. With a similar goal, Held \etal~\cite{held_3d_2025} introduced 3D Convex Splatting, optimizing smooth convex shapes that can stretch asymmetrically to fit corner geometry better than ellipsoids. Alternatively, Kulhanek \etal~\cite{gu_tetrahedron_2024} proposed Tetrahedron Splatting, utilizing tetrahedral primitives to model complex topologies while ensuring continuous volumetric fields suitable for generative tasks.

These recent advancements in primitive design demonstrate that the effectiveness of 3DGS~\cite{Kerbl_2023_3dGS} stems more from the differentiable rendering process itself than from the specific choice of the Gaussian primitive.

\section{Representing real-world scenes}
\label{sec:real_world}
\begin{figure}[t!]
\centering
    \includegraphics[width=\columnwidth]{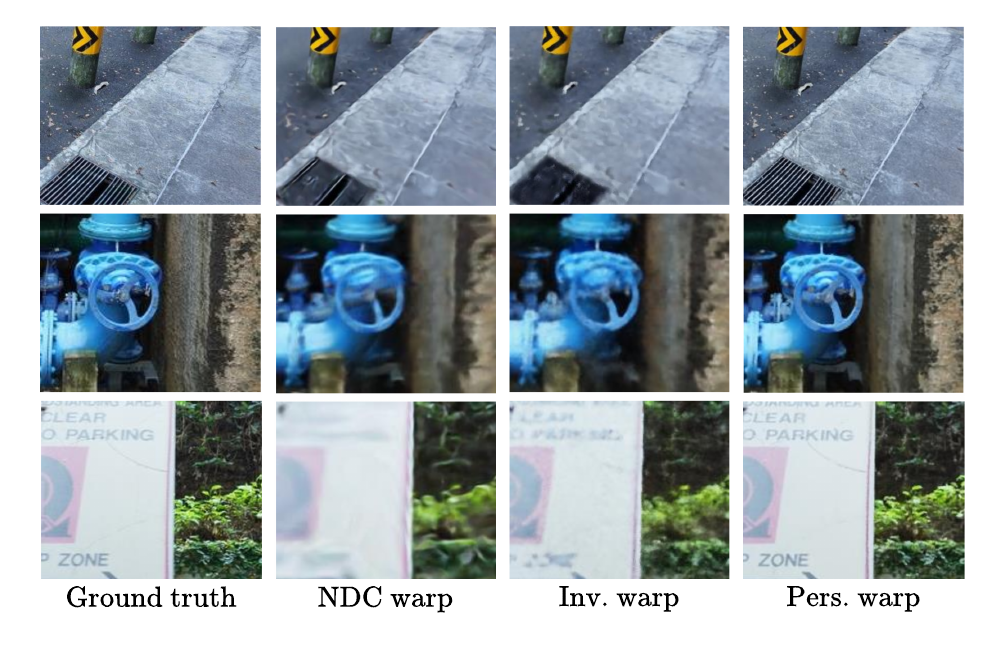}
    \caption{Qualitative comparison of rendered image on the free dataset~\cite{wang2023f} using different warping techniques; results borrowed from~\cite{wang2023f}. On this dataset, inverse warping and perspective warping both outperform classical NDC warping, especially perspective warping, which is able to represent fine details of the tree scenes.}
    \label{fig:unbounded}
\end{figure}
Recent advances in NeRF~\cite{mildenhall2020nerf} and 3DGS~\cite{Kerbl_2023_3dGS} have revolutionized novel view synthesis, achieving photorealistic reconstructions under controlled settings. However, scaling these methods to real-world scenarios remains an open challenge characterized by unconstrained scenes, complex material properties, and sparse input views. 

This section examines the key limitations and emerging solutions across three critical dimensions relevant to real-world deployment.
\First, it reviews methods that extend neural representations to unconstrained or unbounded environments (Section~\ref{sec:unbounded}). \Second, it discusses challenges in modeling scenes with complex material properties and surveys recent approaches designed to address these challenges (Section~\ref{sec:complex_materials}). \textbf{Finally}, it explores techniques for learning accurate radiance field representations from sparse inputs (Section~\ref{sec:gerenalize}).

\subsection{Unconstrained representation}
\label{sec:unbounded}
Standard implicit radiance fields~\cite{mildenhall2020nerf, Kerbl_2023_3dGS} focus on small-scale and object-centric representations. However,  real scenes can be unbounded, geometrically diverse, and contain complex materials often spanning large scales, all introducing additional challenges. This subsection gives an overview of methods developed to adapt neural representations to unbounded scenes (Section~\ref{sec:unbounded_scenes}), unposed images (Section~\ref{sec:unposed}), and large-scale reconstruction (Section~\ref{sec:large_scale_scenes}), using progressive and/or multiscale techniques to address the complexity and scale issues.

\subsubsection{Unbounded scenes}\label{sec:unbounded_scenes}

Applying implicit fields to unbounded scenes, where the camera may point in any direction and content may exist at any distance,  raises two critical issues:
\begin{itemize}
    \item \textbf{Parameterization.} Unbounded 360-degree scenes can occupy an arbitrarily large region of the Euclidean space.
    \item \textbf{Ambiguity.} Content may lie far from the cameras and be observed by few view rays, increasing the inherent ambiguity in reconstructing 3D content from 2D images.
\end{itemize}

\noi A commonly-adopted strategy to address these issues is to use a space-warping method that maps an unbounded space to a bounded space~\cite{barron2022mip,mildenhall2020nerf,zhang2020nerf++}; see Fig~\ref{fig:unbounded}. Three major warping strategies are:

\vspace{6pt}
\textit{(1) Normalized Device Coordinate (NDC) warping.} Used for forward-facing scenes, this technique maps an infinitely far view frustum to a bounded box by squashing the $z$-axis~\cite{mildenhall2020nerf}. While effective for forward-facing scenes, it is unsuitable for $360^\circ$ environments. 

\vspace{6pt}
\textit{(2) Inverse-sphere warping.} For omnidirectional scenes, the inverse-sphere warping can be used to map an infinitely large space to a bounded sphere by the sphere inversion transformation~\cite{zhang2020nerf++,neff2021donerf,barron2022mip}. However, such methods assume special camera trajectory patterns and cannot handle arbitrary ones.

\vspace{6pt}
\noi\textit{(3) Perspective warping.} 
Rather than relying on traditional space warping techniques, such as NDC warping~\cite{mildenhall2020nerf} or inverse sphere warping~\cite{barron2022mip,zhang2020nerf++}, Wang \etal~\cite{wang2023f} propose \emph{perspective warping}. The core idea is to ensure that the axis-aligned grids in the warped space align closely with the camera rays. Given a spatial region $S \subset \rthree$ and a set of visible cameras $\{\camera_i\}_{i=1}^{n_c}$, a warping function $F: \rthree \to \rthree$ is considered proper if, for any two points $\point_1, \point_2 \in S$:
\begin{equation}
    \|F(\point_1) - F(\point_2) \| = \sum_{i=1}^{n_c} \| \camera_i(\point_1)  - \camera_i(\point_2) \|,
\end{equation}
meaning that the distance in the warped space reflects the cumulative projection distances across all visible cameras. 

To handle unbounded spaces with free camera motions, their method (F$^2$-NeRF) subdivides the space using an octree. For each octree node, only the cameras that observe the corresponding region are used to compute the local warping function. Note that while this method gracefully handles arbitrary camera trajectories, it is primarily designed for small-scale scenes.

\subsubsection{Unposed images}\label{sec:unposed}
\begin{figure}[tb]
    \centering
    \includegraphics[trim={0cm 0cm 0cm 1cm}, clip, width=\columnwidth]{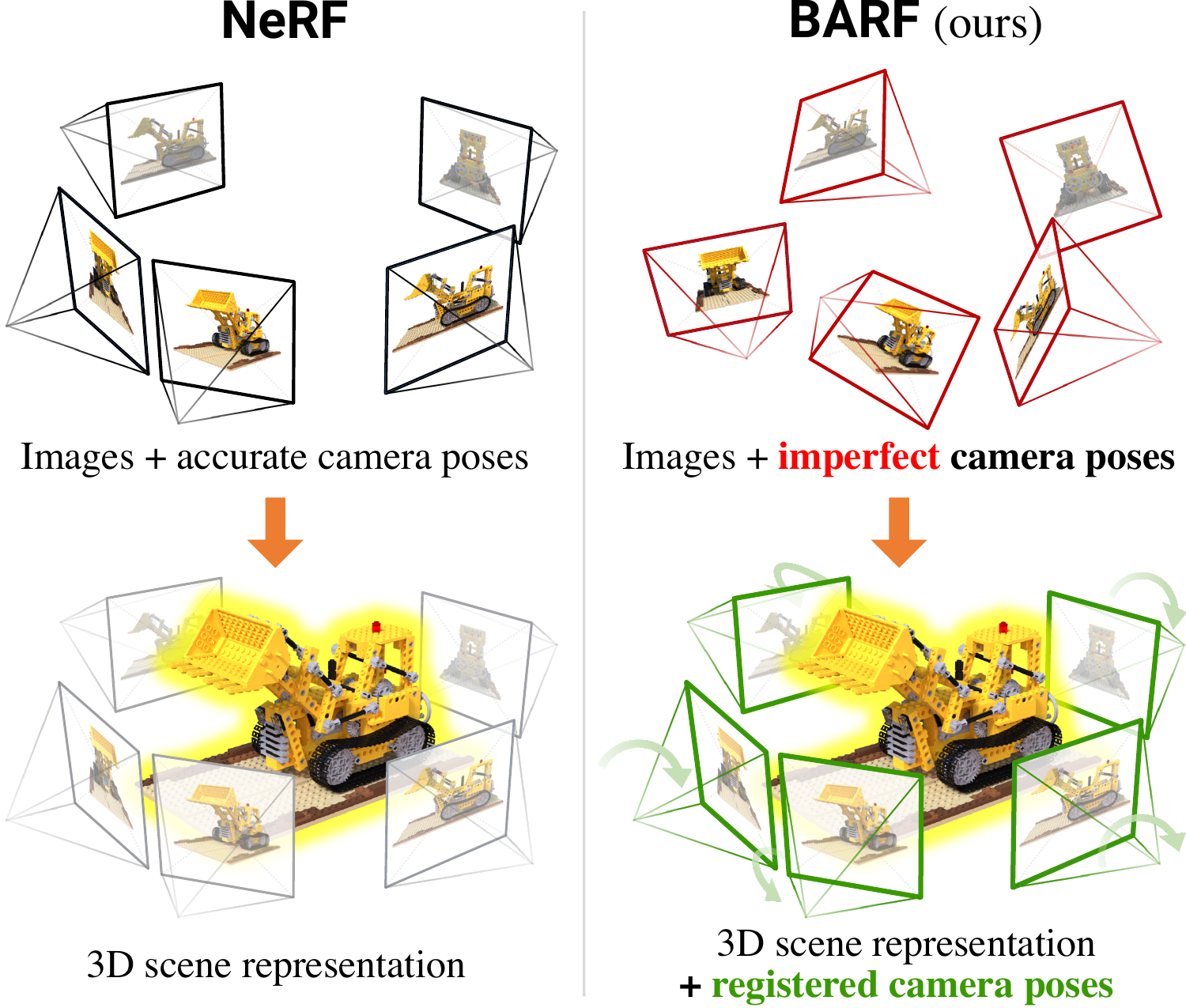}
    \caption{\textbf{Impact of Camera Pose Accuracy.} While standard NeRF heavily relies on perfectly registered camera poses to synthesize sharp scenes, methods like BARF jointly optimize the neural representation alongside imperfect camera poses, relaxing the strict dependency on external Structure-from-Motion pipelines. Image from~\cite{lin2021barf}.}
    \label{fig:joint_optimization}
\end{figure}

Image-based representations such as NeRF~\cite{mildenhall2020nerf} and 3DGS~\cite{Kerbl_2023_3dGS} require accurate camera poses of the input images, typically obtained through Structure-from-Motion (SfM) methods such as  COLMAP~\cite{schonberger2016structure}, for training or at least initialization. This requirement complicates the overall optimization pipeline and limits the potential of these representations for real-world applications. Furthermore, dependence on off-the-shelf pose estimation methods increases inference time and may fail in texture-less regions or when images lack overlap. To eliminate the need for known camera parameters, several works propose to jointly optimize camera poses and neural representation, both in static~\cite{hong2023unifying, wang2021nerfmm, Chen_2023_CVPR, lin2021barf, Shi_2025_nope} and dynamic settings~\cite {liu2023robust}. However, these methods still require rough pose initialization and are limited to small motions. Alternative solutions follow incremental approaches, leveraging mono-depth prior~\cite{bian2022nope}  or temporal continuity~\cite{fu2024colmap} to optimize camera parameters. Several works have extended this idea to obtain generalizable representations~\cite {chen2023dbarf, hong2023unifying, xu2024sparp}, enabling joint pose prediction and representation learning from sparse image collections. 
Ye \etal~\cite{ye2024noposplat} demonstrated that it was possible to efficiently estimate pose and perform novel-view synthesis from unposed images and intrinsic camera parameters using only photometric loss. It is, however, limited to static scenes.

\subsubsection{Large-scale scenes}
\label{sec:large_scale_scenes}

In large, unbounded scenes—\eg city-scale scenes—views range from satellite level, capturing the overview of a scene, to ground-level imagery depicting fine-grained details of a scene. Thus,  changes in imagery can be observed at drastically different scales. This wide span of viewing positions within these scenes requires multi-scale renderings with very different levels of detail, which poses great challenges to neural fields and biases them toward compromised performance. Several methods have been proposed to adapt radiance fields to large scenes by following a divide-and-conquer strategy. 

In Block-NeRF~\cite{tancik2022block} and Mega-NeRF~\cite{turki2022mega}, the scene is divided into blocks, each represented by a small MLP. To increase the efficiency of such methods, the division can be learned~\cite{mi2023switchnerf} and feature encoding can be used~\cite{Xu_2023_CVPR, Song2024City, zhang2025efficient}. 

Inspired by these neural field designs, recent methods have adapted the 3DGS~\cite{Kerbl_2023_3dGS} to represent large-scale scenes. Lin \etal~\cite{Lin_2024_CVPR} introduced a progressive partitioning strategy to divide a large scene into multiple cells where the training cameras and 3D Gaussians are properly distributed with an airspace-aware visibility criterion. Subsequent developments have proposed to reconstruct large-scale scenes following a distributed learning~\cite{yuchen2024dogaussian, gao2024cosurfgs}. Alternatively, methods~\cite{xiangli2022bungeenerf, zhuang2023anti, hierarchicalgaussians24, liu2024citygaussian} proposed Level of Detail (LoD) techniques that regulate the amount of detail needed to represent large-scale scenes. Despite these advances, challenges remain: geometry accuracy is often underexplored, and reconstruction time remains a bottleneck in many practical applications.

\subsection{Material aware representations}
\label{sec:complex_materials}

\begin{figure}[t!]
\centering
    \includegraphics[width=\columnwidth]{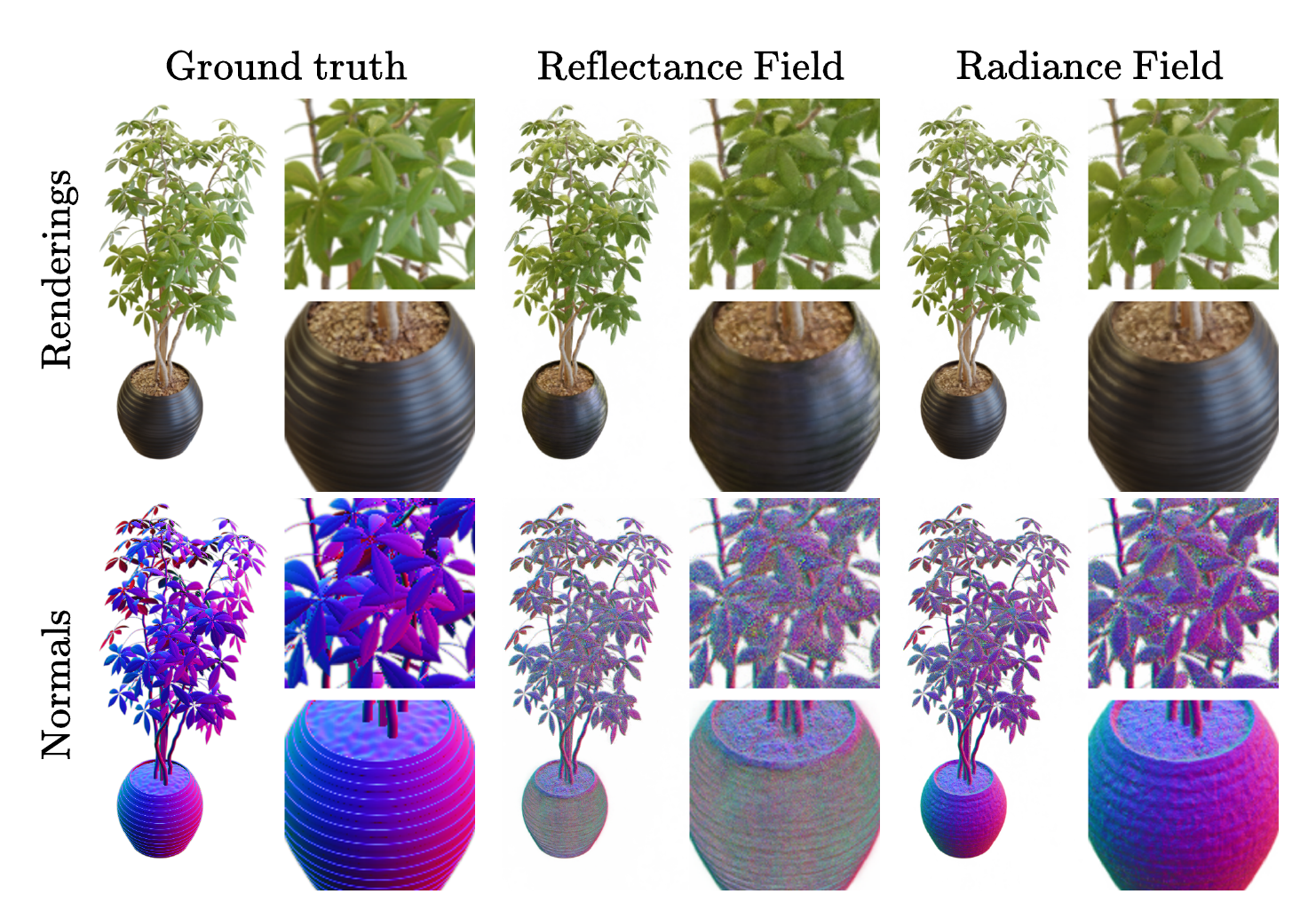}
    \caption{Qualitative comparison of representation capabilities between radiance fields~\cite{barron2021mip} and reflectance fields~\cite{verbin2022ref}. Results, obtained on the NeRF blender dataset~\cite{mildenhall2019local}, borrowed from~\cite{verbin2022ref}}
    \label{fig:complex_materials}
\end{figure}

NeRF~\cite{mildenhall2020nerf} and 3DGS~\cite{Kerbl_2023_3dGS} can generate impressive photo-realistic novel views of a complex 3D scene. However, both methods struggle to produce accurate colors of objects with view-dependent effects. For instance, translucent objects often appear murky, and glossy objects display blurred specular highlights. This is due to the limitations of volume rendering, where features are accumulated in the color space, and to how density values $\volumedensity$—which encode the transmissivity and contribution of a point to a ray's radiance—are computed. This transmissivity should depend on the viewing angles. The accurate rendering of reflective surfaces hinges upon the precise estimation of scene illumination (\eg environment light) and material properties (\eg BRDF), which is the task of inverse rendering. 

Several methods~\cite{verbin2022ref, liu2023nero, srinivasan2021nerv, zhang2023neilf, wang2024nep, Engelhardt_2024_CVPR, lopes2025material} propose using neural fields as 3D representations for inverse rendering and to learn \textit{Reflectance Fields}. In reflectance fields, the scene is represented as a field of volume density, surface normals, and Bidirectional Reflectance Distribution Functions (BRDFs); see Fig~\ref{fig:complex_materials}. This is typically achieved by augmenting NeRF with an additional MLP to model lighting. Subsequent works have improved these representations by modeling BRDFs with spherical Gaussians, by incorporating SDF-based representation in the model, or by introducing regularization to stabilize normals from NeRF’s foggy geometry. Despite these improvements, such methods suffer from extremely time-consuming optimization and slow rendering speed, which limits their viability in practical settings. 

In contrast, recent methods~\cite{jiang2024gaussianshader, liang2024gs, ye20243d} have adapted inverse rendering techniques to 3DGS, simplifying the classical rendering equation to handle specular reflections. These approaches rely on accurate normal estimation, which can be derived from computed depth maps~\cite{liang2024gs} or using the shortest direction axis of Gaussians~\cite{jiang2024gaussianshader, ye20243d}.  The different lighting conditions are then encoded via integrated shading functions. While these 3DGS-based methods achieve rapid convergence,  they still introduce substantial noise in geometry, material, and lighting, which remains a limitation for high-fidelity rendering.

\subsection{Generalizable reconstruction}
\label{sec:gerenalize}

Both vanilla NeRFs~\cite{mildenhall2020nerf} and 3DGS~\cite{Kerbl_2023_3dGS} overfit training data to represent a scene with high quality. Therefore, they only memorize a single scene and can't represent new content without retraining. This limits their use in scenarios requiring zero to few-shot reconstruction and limits the capacity to learn strong knowledge from large datasets. To overcome this structural limit, efforts have been made towards learning generalizable representations. Methods address this by learning a rendering function conditioned on input images, allowing the model to reason about geometry and appearance across different scenes. These methods typically trade the high fidelity of per-scene optimization for the robustness of learned priors. 


Original implicit radiance field representations lack cross-scene generalization capabilities. They typically overlook relations between views and therefore require long per-scene optimization. To mitigate this limitation, methods~\cite{long2022sparseneus} have followed a generalizable approach, aiming to learn a generalizable representation from a large-scale dataset. Such representation can then be fine-tuned and adapted to unseen objects or scenes. This can be achieved by learning generalizable geometric priors from the source images. Such priors are  obtained using feature volumes~\cite{yu2021pixelnerf, ren2023volrecon, long2022sparseneus} or a transformer-based attention module~\cite{liang2024retr, na2024uforecon, zhang2024transplat} to aggregate source views information. 

To further enforce generalization, recent methods leverage deep multiview stereo (MVS). By incorporating MVS strategies into the original NeRF~\cite{mvsnerf} and 3DGS pipeline~\cite{chen2024mvsplat}, they can obtain cross-dataset generalization while learning information about both geometry and appearance. This is typically done by computing cost volumes~\cite{mvsnerf, chen2024mvsplat} that capture cross-view feature matching information or by performing feature matching using an attention mechanism~\cite{chen2024mvsplat}. Unlike standard methods that typically require a dense set of views (e.g., hundreds of images) and primarily rely on individual image features, methods incorporating MVS concepts are designed to work effectively with very sparse inputs (as few as two images). They leverage the geometric cues provided by similarities across different views, significantly improving geometry estimation.

To further scale self-supervised NeRF training, Irshad \etal~\cite{irshad2024nerfmae} proposed using Masked Auto Encoder (MAE) combined with a 3D transformer-based architecture.  This results in a generalized representation that is capable of reconstructing unseen scenes and can be used for transfer learning. Following the same idea, Rajasegaran \etal~\cite{ma_shapesplat_2024} explored the usage of a masked autoencoder to learn a generalizable 3D Gaussian splatting-based representation.

These methods bring generalization capacities to the standard representations. They, however, require significant time on high-performance GPUs (usually several days) to learn the correspondence between 3D geometry and 2D views in advance.

\section{Dynamic scene Representations}
\label{sec:dynamic}

\begin{figure*}[t!]
\centering
    \includegraphics[width=\textwidth]{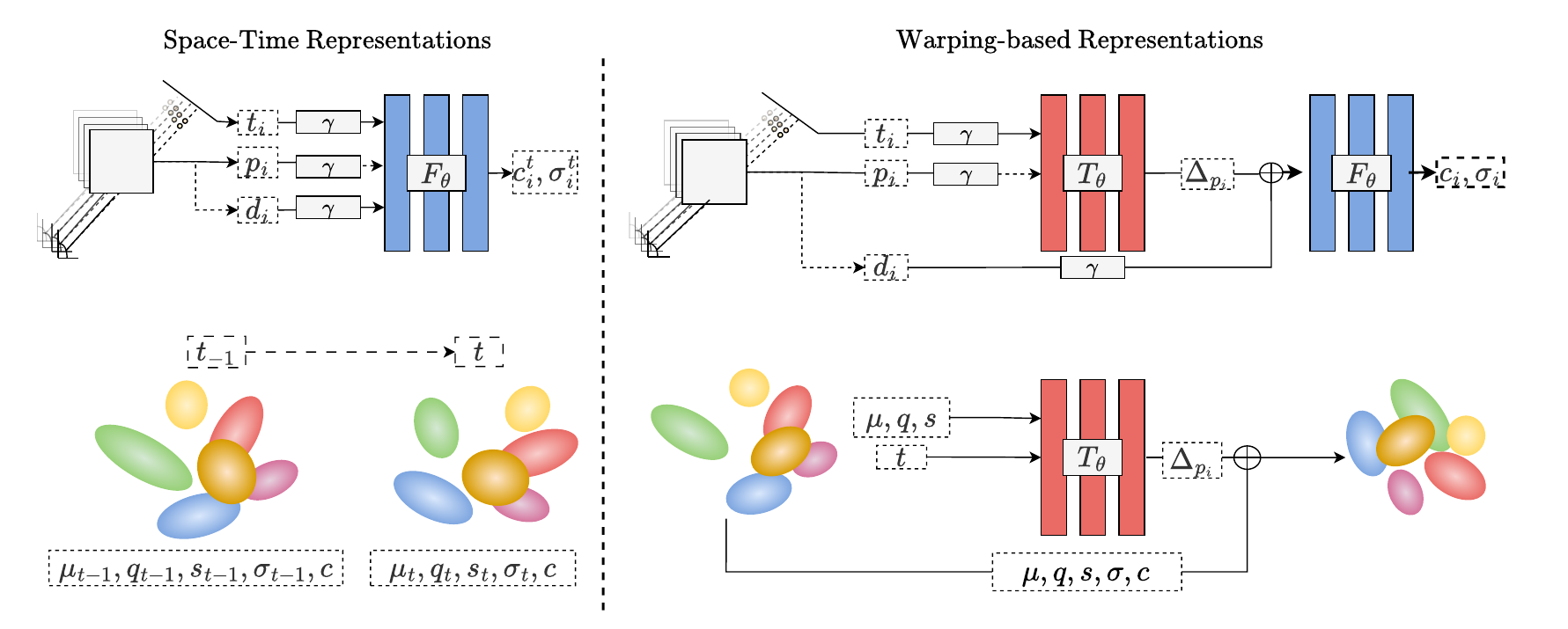}
    \caption{Overview of the two main categories of dynamic representations. (left) \textbf{Space-Time representations} directly condition static 3D representations~\cite{mildenhall2020nerf, Kerbl_2023_3dGS} on time, thereby constructing space-time fields that model temporal changes explicitly. (right) \textbf{Warping-based methods}first learn a deformation field ($F_\theta$) that warps the observed point into a canonical space, and then use a static 3D representations~\cite{mildenhall2020nerf, Kerbl_2023_3dGS} to infer field values for this warped point.}
    \label{fig:dynamic}
\end{figure*}

Real-world 3D scenes are, in most cases, composed of a mix of static and dynamic objects that undergo motions ranging from articulated body movement to soft material bending or foliage fluttering. While 3D representations discussed so far can faithfully capture geometry and appearance, they struggle to represent motion, which limits their practical use in dynamic contexts. 

Motions are typically categorized as rigid or non-rigid. Rigid motion refers to transformations that preserve the internal geometry of the object (\eg, rotation and translation) and can be compactly modeled. This is typically achieved by partitioning the scene into static and rigidly moving components, each modeled with a separate representation that is blended at render time~\cite{lombardi2021mixture, zhang2023nerflets}. In contrast, non-rigid motion involves complex deformations that continuously alter the object's shape over time, such as the articulation of a human body, cloth deformation, or the slight movement of plant leaves. Modeling non-rigid motion remains substantially more challenging, as it requires representations that can account for both spatial and temporal variability. 

This section focuses on methods designed to model both rigid and non-rigid motion, which are classified into two categories:
\begin{itemize}
    \item \textbf{Warping-based methods (Section~\ref{sec:warping_based})} which first learn a global canonical representation and then define a learnable, time-dependent deformation field that warps the canonical space to match the observed scene at each timestep.
    \item \textbf{Space time representations (Section~\ref{sec:space_time_representation})}, which condition the representation with a time component. They model geometry and appearance as a continuous function of both space and time.
\end{itemize}

\subsection{Warping-based Methods}
\label{sec:warping_based}

Warping-based methods~\cite{park2021nerfies,park2021hypernerf,tretschk2021non, lin2024gaussian, wu20244d, yang2024deformable} first learn a temporally global 3D representation, referred to as a canonical representation. The motion is then modeled by a deformation field, which can be used to warp all the 3D points in the canonical space to their deformed state at each time step. This automatically ensures long-term correspondence, allowing consistent novel view synthesis and smooth interpolation in the time dimension. 
In practice, warping-based methods can be expressed as a combination of two functions: a deformation network $\transformation$, which maps a point $\point$ to its canonical counterpart $\point'$, and a static representation $\implicitfunc$, which infers the implicit field from $\point'$. This can be formalized as:
\begin{equation}
    \begin{aligned}
        \neuralsurfacefunc =& \implicitfunc \circ \transformation \text{ with }\\
        \transformation : & (\point, \motionconditions) \to \point' \text{ and }\ \implicitfunc:\point'[, \viewdir; \conditions] \to \shapeproperties,
    \end{aligned}
    \label{eq:warp_method}
\end{equation}

\noi where the operator $\circ$ denotes function composition. 

Methods in this class differ based on how the deformation field $\transformation$ is modeled. Some methods learn a backward warp that represents a mapping of deformed states back to the canonical representation (as in Eq~\ref{eq:warp_method}). However, when attempting to map 3D points back to the canonical space across different time steps, the backward warp often results in discontinuous mapping, as various object configurations can occupy the same spatial location. In contrast, learning a forward warp~\cite{guo2023forward} mitigates these limitations by providing a smooth and continuous deformation mapping from the canonical space to dynamic time steps.

In general, the deformation field is either modeled by an MLP~\cite{liang2024gaufregaussiandeformationfields, huang2024sc, yang2024deformable} or explicitly using polynomial~\cite{lin2024gaussian, Li_2024_CVPR_spacetime}, Fourier transform~\cite{lin2024gaussian}, or a learned basis function~\cite{kratimenos2024dynmf}.  

Despite their flexibility, warp-based methods struggle with topology changes, limiting their applicability primarily to object-level modeling, where the object maintains a consistent structure over time.

\subsection{Space-time Representations}
\label{sec:space_time_representation}

Space-time representations aim to model dynamic scenes by extending the query space of Eq.~\ref{eq:3d_representation}. The query $\mathbf{q}$ is expanded to include time, becoming $\mathbf{q} = (\point, \viewdir, \thetime)$, yielding the function:
\begin{equation}
     \neuralsurfacefunc_\params: \big(\gamma(\point, \viewdir, \thetime)\big) \to \shapeproperties.
\end{equation}

Methods in this class omit the deformation network $\transformation$ and instead represent the dynamic scene using a single unified function that directly regresses dynamic properties. In practice, this is achieved by directly conditioning a 3D representation with a time component. As illustrated in Fig~\ref{fig:dynamic}, this representation may take the form of:
\begin{itemize}
    \item A neural implicit model conditioned on a time variable~\cite{Gao2021_DynamicViewNeRF, li2021neural, xian2021space}
    \item A set of 3D Gaussians shared across frames, where time-dependent parameters encode motion~\cite{luiten2024dynamic}.
\end{itemize}

Because space-time representations impose fewer geometry constraints than warp-based methods, they can model a larger range of motion and
are tolerant of topology changes. However, this flexibility leads to time inconsistency and makes the representation dependent on the number of observations available. Time consistency can be enforced locally via the usage of flow fields~\cite{li2021neural, du2021neural, liu2023robust, Li_2023_CVPR}, which defines forward and backward mappings between neighboring time steps when available. 

Alternatively, instead of modeling $ \neuralsurfacefunc$ as a function of time, one can define the encoding $\enc$ as a function of time~\cite{park2023temporal}. In other words:
\begin{equation}
    \neuralsurfacefunc: (\enc(\point, \thetime), [\enc(\viewdir)],) \to \shapeproperties.
\end{equation}

\noi  The encoding function $\enc$ can be either an MLP-based neural network or an explicit feature grid. In practice, the time interval is divided into equally spaced time slots, and the MLPs compute features at these discrete time slots. The feature of a given point $\point\in \rthree$ at any time $\thetime$ is computed by interpolating between neighboring time slots. This strategy decreases the computational burden and enables faster training.

\subsection{Feature-grid encoding for dynamic representations}
\label{sec:grid_based_dynamic_Nerf}

Both warping-based and space-time-based dynamic representations generally use MLPs to represent dynamic scenes. This requires extensive computation time for both training and rendering. Recent works focusing on extending feature grid-based optimization~\cite{muller2022instant, chen2022tensorf, chan2022efficient}, which have proven effective for accelerating static 3D representations, to dynamic content. A direct way to represent dynamic content is to build upon a static representation and make it time-dependent. This has been applied to voxel-grid~\cite{fang2022fast, song2023nerfplayer}, Plenoctrees~\cite{wang2022fourier}, triplane~\cite{shao2023tensor4d, cao2023hexplane, fridovich2023k} or more recently with wavelet-based representations~\cite{lou2024darenerf}. However, directly using such feature encoding for each time step results in substantial memory costs, making it impractical for long sequences or high temporal resolution. One way to address this issue is to use a single canonical feature space, as proposed in~\cite {liu2023robust, lin2024ced}. In this setup, an input point $\point$ is first deformed to its canonical space, where its corresponding feature is queried. An alternative solution is to use representations based on a small number of space-time feature planes combined with lightweight MLPs~\cite{cao2023hexplane, shao2023tensor4d, fridovich2023k}.

\section{Applications}
\label{sec:applications}

\begin{figure}[t!]
  \centering
  \begin{subfigure}[b]{\columnwidth}
    \centering
    \includegraphics[width=0.9\columnwidth]{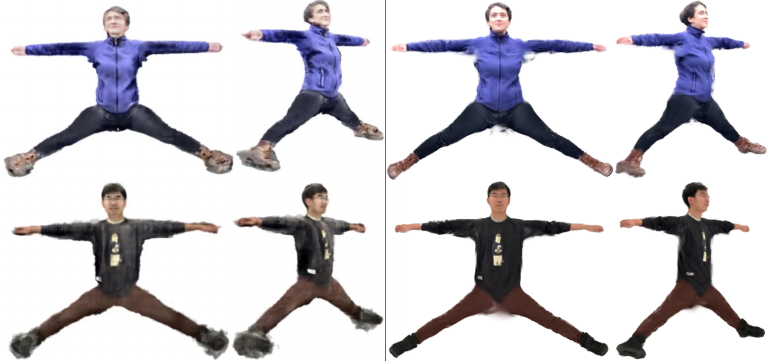}
    \caption{}
    \label{fig:human_reconstruction}
  \end{subfigure}

  \begin{subfigure}[b]{0.48\columnwidth}
    \centering
    \includegraphics[width=\textwidth, trim=0 680 940 0, clip]{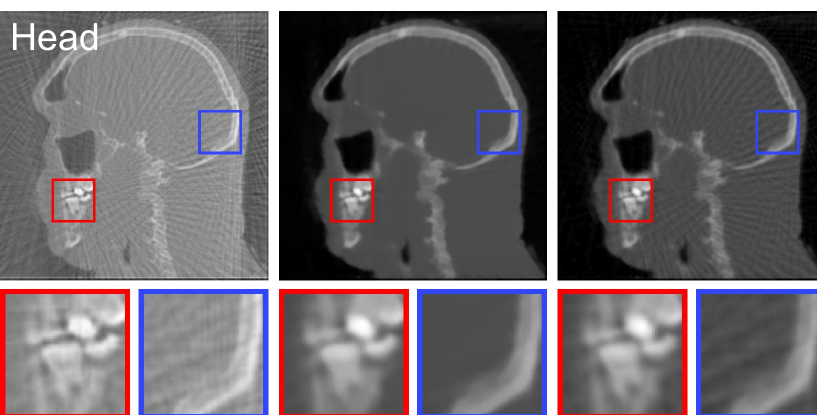}
    \caption{}
    \label{fig:medical}
  \end{subfigure}
  \hfill
  \begin{subfigure}[b]{0.48\columnwidth}
    \centering
    \includegraphics[width=\textwidth, trim=50 0 0 30, clip]{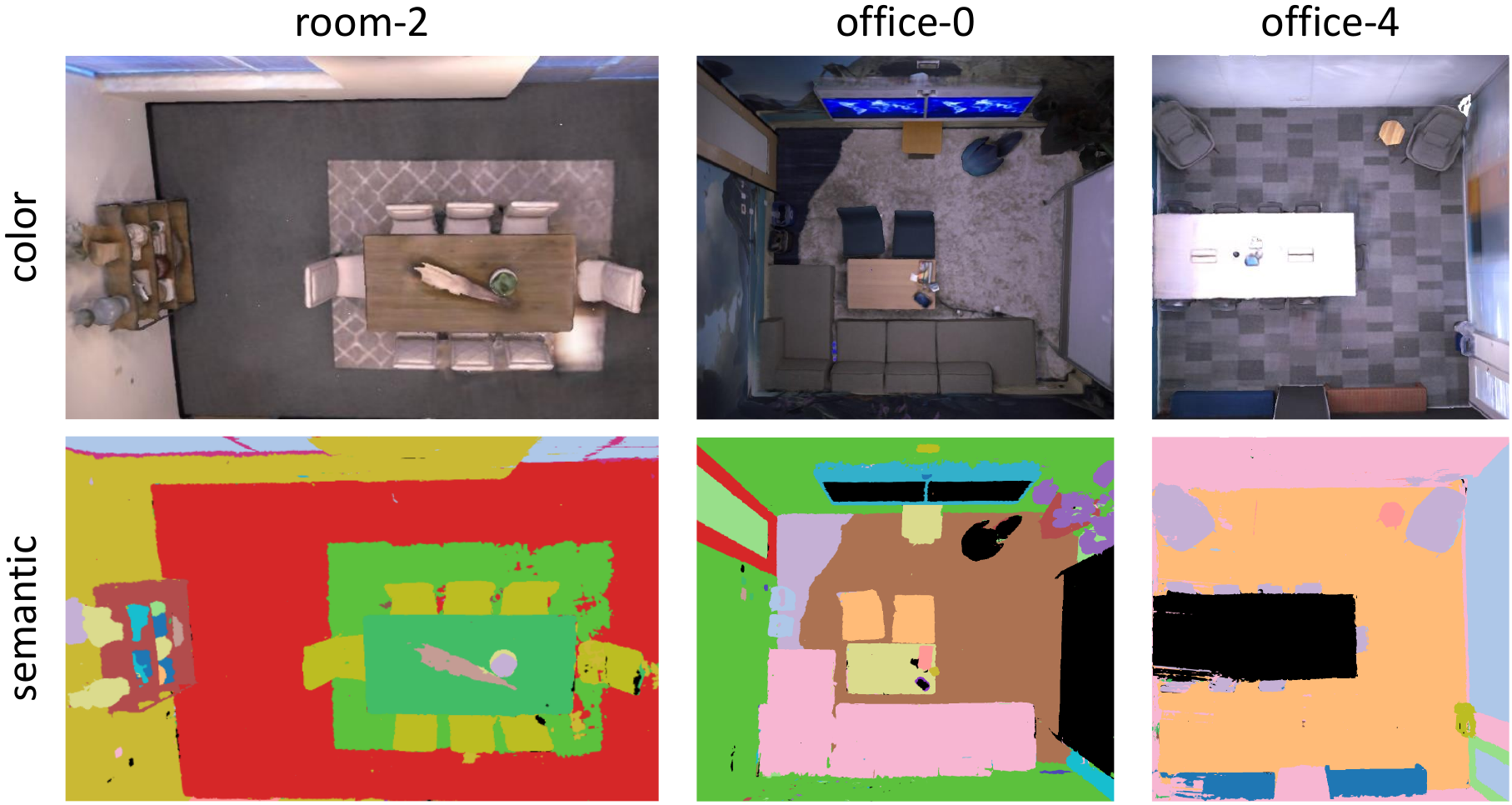}
    \caption{}
    \label{fig:medical_imaging}
  \end{subfigure}

  \begin{subfigure}[b]{0.48\columnwidth}
    \centering
    \includegraphics[width=\columnwidth, trim=0 80 215 0, clip]{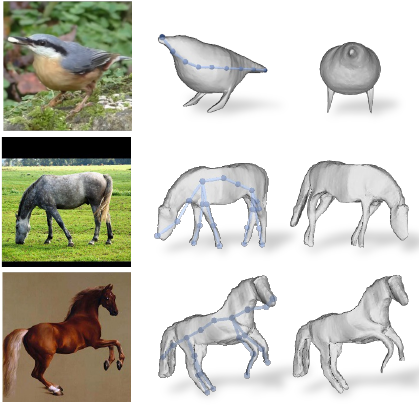}
    \caption{}
    \label{fig:animal}
  \end{subfigure}
  \hfill
  \begin{subfigure}[b]{0.48\columnwidth}
    \centering
    \includegraphics[width=0.8\columnwidth]{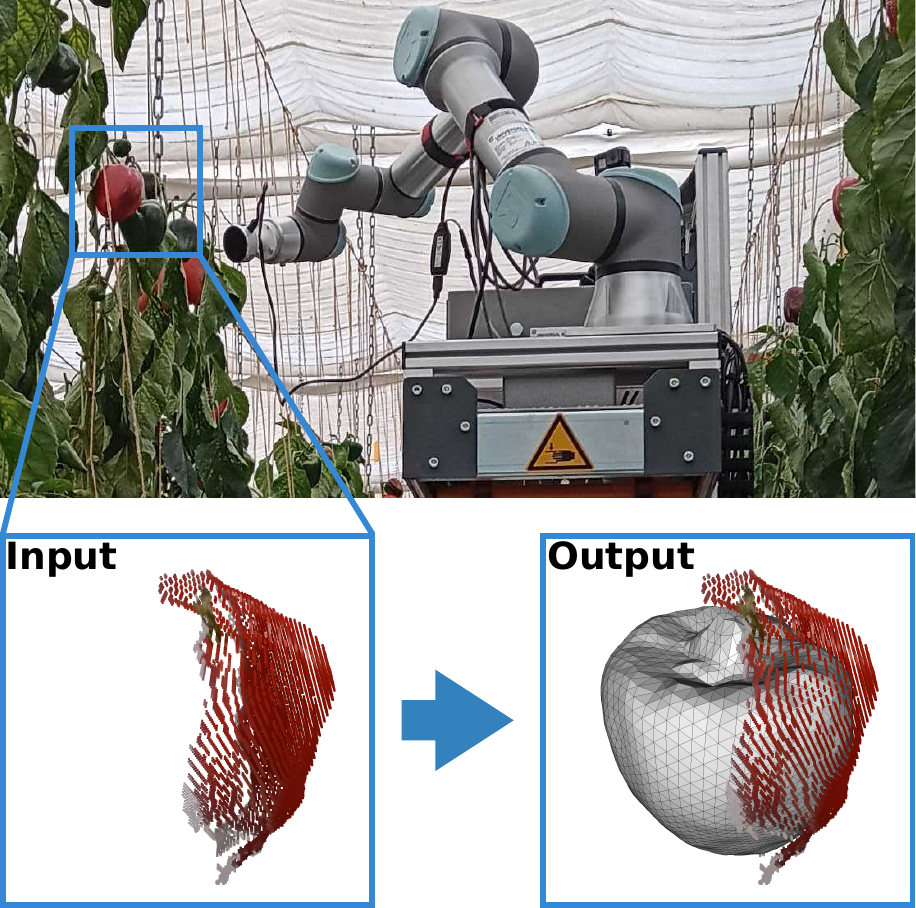}
    \caption{}
    \label{fig:animal_reconstruction}
  \end{subfigure}

  \caption{Composite figure illustrating various application examples of neural 3D representations. (a) Human reconstruction. (b) Medical imaging. (c) SLAM (Simultaneous Localization and Mapping). (d) Animal reconstruction. (e) Plant reconstruction. Images adapted from~\cite{wu2023magicpony, magistri2022robotics, sax_nerf, kocabas2024hugs,zhu2024sni}.}
  \label{fig:applications}
\end{figure}

3D reconstruction has traditionally been both time- and resource-intensive. Early methods either required access to expensive 3D hardware or were limited to highly controlled environments, making them accessible only to fields with substantial resources. Recent advancements in 3D learned representations have addressed these limitations, enabling broader applicability across a wide range of scientific and practical domains. This paradigm shift significantly lowers the barrier to entry for 3D vision and makes it accessible to more research and application areas. As shown in Figure \ref{fig:applications}, modern representations~\cite{mildenhall2020nerf, Kerbl_2023_3dGS} have expanded into diverse scientific fields. This deployment has revealed that generic "vanilla" representations often fail to satisfy specific real-world constraints, such as anatomical plausibility, real-time adaptability, or controllability. Hence, domain-specific methods introduced representation enhancement, becoming active drivers of the domain. While the range of potential applications is vast, this section focuses on three key domains—Biological (Section~\ref{sec:humans}), Robotics (Section~\ref{sec:robotics}), and Generative Content (Section~\ref{sec:generative})—selected because they clearly demonstrate how domain-specific constraints force fundamental modifications to the underlying 3D representation.

\subsection{Biological and Digital Humans}
\label{sec:humans}

Reconstructing humans~\cite{liu2021neural,hu2024gaussianavatar,kocabas2024hugs} and animals~\cite{badger20203d, Ruegg_2023_CVPR, wu2023dove} imposes strict anatomical and kinematic constraints. Unlike static scenes, biological subjects deform non-rigidly and must adhere to skeletal structures. Generic representations often struggle with these articulated motions, producing artifacts or physically impossible distortions.
To address this, recent methods enhance the representation by embedding kinematic priors. By anchoring neural fields or Gaussian splats to parametric templates (e.g., SMPL for humans~\cite{loper2015smpl}, MANO for hands~\cite{MANO_2017}, 3DMM~\cite{booth20163d} for the face), the representation is forced to respect joint limits and skinning weights. This hybrid approach—combining learnable volumetric appearance with explicit skeletal guidance—has become the standard for high-fidelity digital avatars and motion capture~\cite{liu2021neural, Jiang2022Neuman}.

\subsection{Multi-Modal Adaptability}
\label{sec:multimodal}

Beyond standard RGB imagery, modern 3D representations have demonstrated remarkable flexibility in modeling diverse physical modalities, unlocking new capabilities in scientific and industrial domains. In medical imaging, recent works have adapted both neural representations and primitive-based methods to reconstruct high-fidelity anatomical structures from sparse X-ray projections~\cite{corona2022mednerf, Chen_2023_ICCV,cai2024xray3dgs} and to synthesize novel ultrasound views by integrating acoustic propagation models~\cite{pmlr-v227-wysocki24a}.
In robotics and autonomous navigation, representations are increasingly tailored to non-visual sensors; methods~\cite{Huang_2023_ICCV,tao2024lidar,zhang2024nerf,mcdermott2025probabilistic} have adapted neural representations for LiDAR by formulating a differentiable framework for novel LiDAR view synthesis~\cite{tao2024lidar}, introducing a beam spreading model~\cite{Huang_2023_ICCV}, and incorporating probabilistic losses~\cite{mcdermott2025probabilistic} into neural representations to handle the stochastic nature of laser returns. This cross-modal adaptability transforms 3D representations from purely visual tools into versatile substrates for fusing heterogeneous sensor data.

\subsection{Robotics and Situated AI}
\label{sec:robotics}

Robotics and autonomous navigation impose the strictest constraints on 3D representations: they must be learnable \textit{incrementally} (as the robot explores), \textit{in real-time} (for loop closure), and robust to sensor noise. While early neural SLAM methods~\cite{sucar2021imap,zhu2022nice} adapted implicit neural representations for online use, they were often limited by the slow convergence of MLPs. Recent methods accelerated this capability by adopting primitive splatting~\cite{Keetha_2024_CVPR,Matsuki_2024_CVPR} for SLAM. They demonstrate that explicit Gaussians can be inserted, moved, or deleted instantaneously, enabling real-time, a critical requirement for agile robots. Beyond 2D or 3D Slam, this pipeline can be augmented to 4D tracking and mapping of points using dynamic 3D~\cite{seidenschwarz2025dynomo} or 2D Gaussians~\cite{Matsuki_2025_CVPR}.

\subsection{Generative Content Creation}
\label{sec:generative}
In creative industries (gaming, VR/AR, VFX), the primary goal shifts from reconstructing \textit{what is observed} to generating \textit{what is imagined}. Vanilla 3D representations, which rely on multiview consistency, cannot inherently hallucinate plausible geometry or texture for unseen regions. To address this, recent methods enhance 3D representations by infusing them with semantic and visual priors distilled from Internet-scale 2D foundation models~\cite{radford2021learning, Rombach_2022_CVPR}. This integration transforms the 3D representation from a purely geometric container into a semantically aware, generative medium.

\vspace{6pt}
\noi\textbf{Text-to-3D via Score Distillation.}
The breakthrough in this domain is the ability to optimize a 3D representation (NeRF or 3DGS) using a frozen 2D text-to-image diffusion model~\cite{radford2021learning}. This is typically achieved via Score Distillation Sampling (SDS), a technique where parameters are updated until the rendered views match a text prompt. Methods like Magic3D~\cite{lin2023magic3d} and Fantasia3D~\cite{chen2023fantasia3d} use this to generate high-fidelity 3D assets from scratch. Recent adaptations to primitive-based splatting~\cite{yi2023gaussiandreamer, chen2024text, lin2025diffsplat} have accelerated this process from hours to minutes, enabling iterative, real-time creative workflows that are essential for artists.

\vspace{6pt}
\noi\textbf{Single-Image-to-3D.}
Generating a full 3D asset from a single image requires strong priors to resolve extreme occlusion. Generative 3D methods leverage 2D diffusion models that are fine-tuned to be multiview-aware~\cite{liu2023one2345, liu2024syncdreamer}. These models predict consistent novel views (or normal maps) from the single input, which are then fused into a consistent 3D representation.

\vspace{6pt}
\noi\textbf{Semantic Awareness and Editing.}
Beyond generating geometry, creative applications require understanding \textit{what} an object is to enable editing. Generic representations are semantically blind. To fix this, methods like LERF~\cite{Kerr_2023_ICCV} and LangSplat~\cite{Qin_2024_CVPR} distill high-dimensional language embeddings (e.g., from CLIP~\cite{radford2021learning}) directly into the volumetric field. This enhances the representation into a queryable, semantically aware space, allowing users to select, recolor, or delete object parts using natural language commands (e.g., "remove the red car").

\section{Future research and conclusion}
\label{sec:future_research}
Recent years have seen remarkable progress in 3D representations, while traditional representations (\ie meshes, point clouds, voxels) still dominate many applications. The rise of learning-based 3D representations like neural radiance fields or primitive-based methods has transformed the field. They have introduced new paradigms for modeling complex geometry and appearance, achieving unprecedented levels of realism. However, despite this success, several challenges remain, and the field continues to evolve rapidly. This section outlines key limitations in current methods and highlights open research directions that could shape the future of 3D learning. In doing so, it provides a foundation for future innovation and broader deployment across application domains.

\vspace{6pt}
\noi\textbf{Computational vs. Memory Efficiency Trade-off.} 
A major bottleneck in choosing the right 3D representation is the inherent trade-off between computational speed and memory usage. Implicit neural representations are memory-efficient because they only require a set of network weights to represent 3D. However, they suffer from prohibitive training and rendering times due to the need for millions of MLP evaluations at inference. While the recent rise of primitive-based methods has drastically accelerated rendering by removing the MLP, it does so at the cost of increased memory usage. For such 3D representations to become viable across scientific and industrial domains, it is essential to find an effective trade-off between computational efficiency and memory costs. To mitigate this trade-off, recent research has pivoted toward hybrid representations that combine neural representations with explicit processes. Early attempts incorporating discretization (e.g., hash grids~\cite{muller2022instant}, triplanes~\cite{chan2022efficient}, sparse grids~\cite{fridovich2022plenoxels}) successfully accelerated NeRFs but often reintroduced complexity or resolution limits. Another promising direction is to embed neural expressivity directly into splattable primitives. For instance, Zhou \etal~\cite{zhou_splat_2025} introduces \textit{splattable neural primitives}, bounded volumes parameterized by lightweight MLPs that can be rendered via analytical integration and splatting. This approach matches the real-time speed of 3DGS while reducing primitive counts by an order of magnitude, effectively solving the memory-efficiency limit. We believe that this line of work will spur further hybrid representations that bridge neural implicit fields and primitive-based splatting, and that it constitutes a promising direction for future research.

\vspace{6pt}
\noi\textbf{From Optimization to Amortized Inference.}
Most current success stories rely on per-scene optimization (overfitting), which is slow and fragile. The "Generalization" problem is rapidly shifting toward Large Reconstruction Models (LRMs)~\cite{hong_lrm_2023,tang_lgm_2025}. Future research will likely move away from conditioning NeRFs on local features~\cite{yu2021pixelnerf} toward leveraging pre-trained foundation models. Emerging approaches like DUSt3R~\cite{Wang_2024_CVPR}, Mast3r~\cite{leroy2024grounding} or VGGT~\cite{wang_vggt_2025} demonstrate that 3D structure can be regressed directly from images in a feed-forward pass, bypassing traditional Structure-from-Motion entirely. Developing representations that act as efficient "decoders" for these foundation models—supporting zero-shot reconstruction without the need for minutes of optimization—is a critical open frontier.

\vspace{6pt}
\noi\textbf{Robustness and practical deployment.}
Many pipelines rely on fragile prerequisites (accurate camera poses, sufficient overlap, static scenes, controlled exposure/lighting). For instance, obtaining reliable poses via SfM (e.g., COLMAP~\cite{schonberger2016structure}) is often a major failure point in texture-less areas or sparse captures; jointly optimizing pose and representation improves usability but remains sensitive to initialization and motion~\cite{lin2021barf, Chen_2023_CVPR, hong2023unifying, ye2024noposplat}. Beyond pose, real deployments face rolling shutter, motion blur, changing illumination, dynamic content, and sensor mismatch (RGB/RGB-D/LiDAR). A key open problem is to design representations and objectives that degrade gracefully under these violations, and to reduce pipeline complexity (fewer external stages, fewer hyperparameters), which is a concrete barrier to adoption outside research settings.

\vspace{6pt}
\noindent\textbf{Resolving the Shape-Radiance Ambiguity.}
A fundamental open challenge in learning-based 3D vision is the inherent tension between rendering quality (\textit{photometric consistency}) and surface accuracy (\textit{geometric fidelity}). Current state-of-the-art methods, such as NeRF~\cite{mildenhall2020nerf} and 3DGS~\cite{Kerbl_2023_3dGS}, are predominantly optimized to minimize photometric error. Consequently, they tend to solve the objective by generating volumetric artifacts—such as "fog," "floaters," or view-dependent billboards—that satisfy the rendering loss from training angles but fail to represent valid physical structures. While this \textit{Shape-Radiance Ambiguity} is acceptable for pure Novel-View Synthesis (NVS), it severely restricts applicability in geometry-critical domains such as metrology, robotic interaction, and physics simulation. Recent approaches~\cite{li2023neuralangelo,guedon2024sugar,Huang2DGS2024} attempt to bridge this gap by incorporating geometric regularization or surface-aligned primitives. However, obtaining a representation that is simultaneously photorealistic, geometrically accurate, and computationally efficient remains an unsolved problem. This limitation explains the continued dominance of traditional explicit representations in engineering workflows and highlights a critical direction for future research: developing unified representations that do not compromise geometry for the sake of visual quality.

\vspace{6pt}
\noi\textbf{Physical plausibility and real-world constraints.}
Current neural representations are largely disconnected from physical principles. While radiance fields are excellent at photorealistic rendering, they typically ignore material properties (\eg reflectance, transparency) and dynamic interactions (\eg collisions, fluid flow). This limits their utility in domains such as robotics, simulation, and scientific modeling. Recent works~\cite{Zhai_2024_CVPR} proposed to integrate physically-based rendering (PBR) into implicit representations, enabling material editing. Others have coupled volumetric fields with physical knowledge for dynamic simulation~\cite{feng2023pienerf,xie2023physgaussian,le2023differentiable, ni2024phyrecon}. However, these methods remain niche and struggle with generalization. A key direction for future work is the integration of universal physical priors into learned 3D, paving the way for use in predictive simulations, robotic manipulation, and interactive virtual environments.

\vspace{6pt}
\noi
\textbf{Conclusion.} In summary, the field of 3D and 4D representations has made significant strides over the past few years. Yet important challenges and opportunities remain—particularly in the areas of efficiency, adoption, and physical realism. Future research should aim to develop representations that generalize across diverse scenes and object types, support efficient training and real-time inference, maybe through the usage of hybrid representations, combining the speed of primitives with the compactness of neural networks. They must also incorporate semantic and physical understanding to move beyond purely visual reconstruction.
Addressing these challenges will not only enhance the accuracy and applicability of 3D reconstruction techniques but can also expand their utility across robotics, medicine, simulation, and digital content creation. As the field continues to develop, attention is gradually moving away from isolated technical advancements toward more integrated systems—ones that can handle real-world complexity, understand physical interactions, and work smoothly with both 2D and 3D sensing technologies. The intersection of machine learning, computer vision, and computer graphics is set to drive the next wave of breakthroughs in this space.

\section*{Ethical Approval}
This article is a survey of existing work on 3D representations. 
It does not involve specific data collection, and therefore, no ethical approval was required.

\section*{Acknowledgment}

This work is based upon work supported by the ANRT (Association nationale de la recherche et de la technologie) in France with a CIFRE fellowship granted to DOWNS, and is supported in part by the Australian Research Council (ARC) Discovery Project no. DP220102197 and ARC Future Fellowship no. FT250100448.

\bibliographystyle{template/eg-alpha-doi} 
\bibliography{reconstruction}

\end{document}